\pdfoutput=1

\documentclass[11pt]{article}

\usepackage{acl}

\usepackage{times}
\usepackage{latexsym}

\usepackage[T1]{fontenc}

\usepackage[utf8]{inputenc}

\usepackage{microtype}

\usepackage{amsmath}
\usepackage{booktabs}
\usepackage{amssymb}
\usepackage{graphicx}
\usepackage{color, colortbl}
\usepackage{enumitem}
\usepackage{xspace}
\usepackage{caption}
\usepackage{subcaption}
\usepackage{breakurl}

\usepackage{minitoc}

\usepackage{booktabs}       
\usepackage{amsfonts}       
\usepackage{nicefrac}       
\usepackage{microtype}      
\usepackage{xcolor}         

\usepackage{graphicx}
\usepackage{algorithm}
\usepackage{algorithmic}
\usepackage{amsmath}
\usepackage{subcaption}
\usepackage{multirow,tabularx}
\usepackage{makecell}

\usepackage{xcolor}
\usepackage{natbib}

\newcommand{\ours}[1]{TIP}
\newcommand{\method}[1]{Text-Image Prompting}
\newcommand{\uplan}[1]{GPT3 + Stable-Diffusion}
\newcommand{\uplaninstruct}[1]{InstructGPT + Stable-Diffusion}
\newcommand{\mplan}[1]{TIP (Ours)}
\newcommand{\mplanwoti}[1]{TIP (w.o. T2I Bridge)}
\newcommand{\mplanwoit}[1]{TIP (w.o. I2T Bridge)}
\newcommand{\tib}[1]{Text-to-Image Bridge}
\newcommand{\tti}[1]{T2I-B}
\newcommand{\itb}[1]{Image-to-Text Bridge}
\newcommand{\itt}[1]{I2T-B}
\newcommand{\dthree}[1]{Text-Davinci-003}
\newcommand{\dtwo}[1]{Text-Davinci-002}
\newcommand{\vgt}[1]{Image Ref + OFA-Caption}
\newcommand{\vgtblip}[1]{Image Ref + BLIP-Caption}
\newcommand{\tgt}[1]{Text Ref + Stable-Diffusion}
\newcommand{\tgtdalle}[1]{Text Ref + DALLE}
\newcommand{\uthree}[1]{Text-Davinci-003 + Stable-Diffusion}
\newcommand{\utwo}[1]{Text-Davinci-002 + Stable-Diffusion}
\newcommand{\davinci}[1]{Davinci}
\newcommand{\ada}[1]{Text-Ada}
\newcommand{\step}[1]{Text-Davinci-003 (Step-based) + Stable-Diffusion}
\newcommand{\wiki}[1]{\textsc{WikiPlan}}
\newcommand{\recipe}[1]{\textsc{RecipePlan}}
\newcommand{\wikihow}[1]{\textsc{WikiHow}}
\newcommand{\recipeqa}[1]{\textsc{RecipeQA}}

\definecolor{my_green}{RGB}{34,139,34}

\newcommand{\blue}[1]{\textcolor{blue}{#1}}
\newcommand{\green}[1]{\textcolor{my_green}{#1}}
\newcommand{\red}[1]{\textcolor{red}{#1}}
\newcommand{\pink}[1]{\textcolor{pink}{#1}}

\title{Multimodal Procedural Planning via Dual Text-Image Prompting
}

\author{Yujie Lu$^1$, Pan Lu$^2$, Zhiyu Chen$^1$, Wanrong Zhu$^1$, Xin Eric Wang$^3$\\\textbf{William Yang Wang$^1$}\\
$^1$University of California, Santa Barbara, CA, USA\\ 
\texttt{\{yujielu,zhiyuchen,wanrongzhu,wangwilliamyang\}@ucsb.edu}\\
$^2$University of California, Los Angeles, CA, USA\\
\texttt{lupantech@gmail.com}\\
$^3$University of California, Santa Cruz, CA, USA\\
\texttt{xwang366@ucsc.edu}\\
}
\begin{document}
\twocolumn[{%
\renewcommand\twocolumn[1][]{#1}%
\maketitle

\begin{center}
    \includegraphics[width=0.98\linewidth]{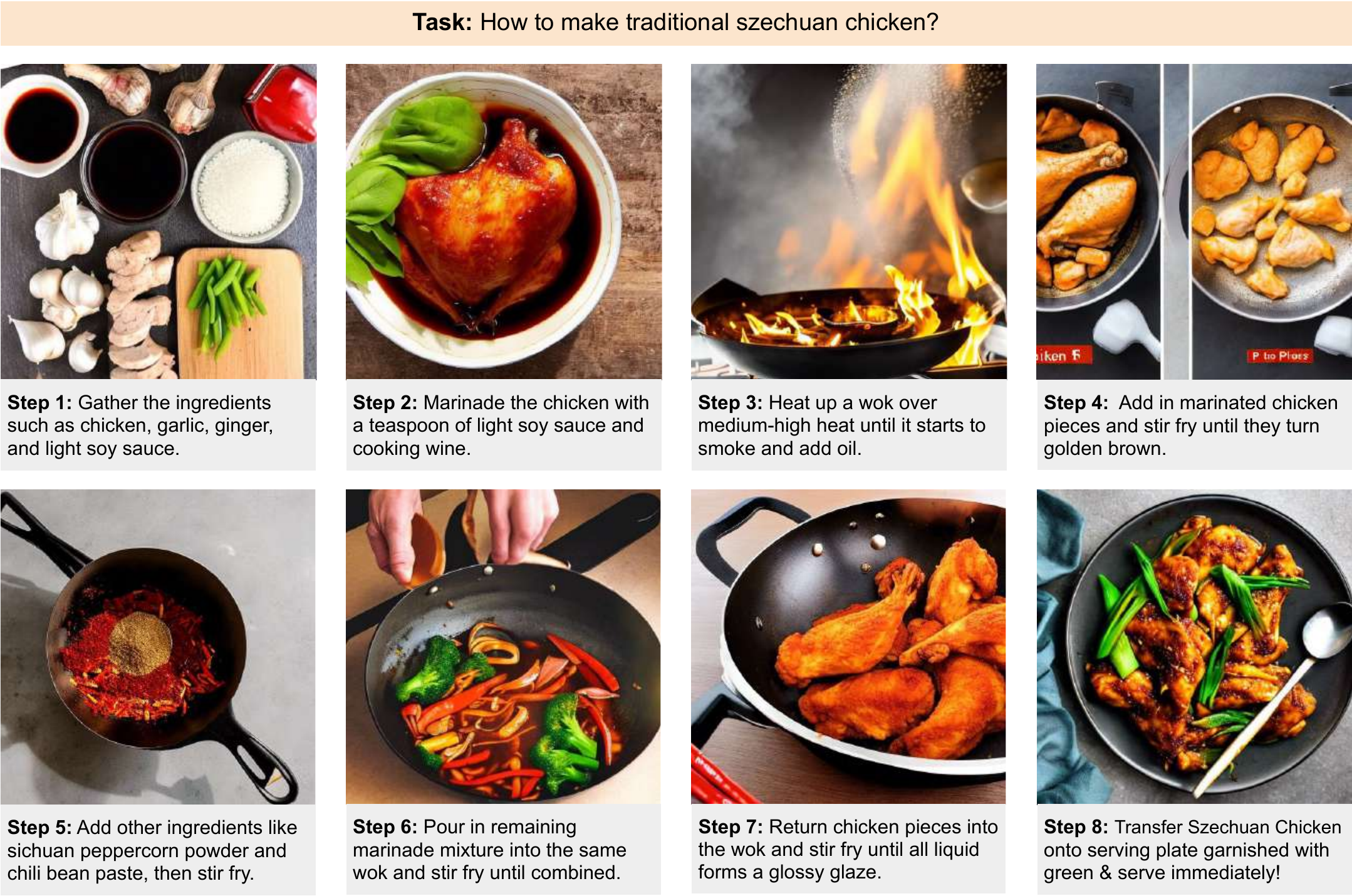}%
    \vspace{-1.5mm}
    \captionof{figure}{
    Our dual Text-Image Prompting (TIP) model generates coherent and authentic \textit{multimodal procedural plans} with multiple steps towards a high-level goal, providing useful guidelines in task completion.
    }
    \label{fig:teaser}
    \vspace{2.7mm}
\end{center}%
}]

\begin{abstract}
Embodied agents have achieved prominent performance in following human instructions to complete tasks. However, the potential of providing instructions informed by texts and images to assist humans in completing tasks remains underexplored.
To uncover this capability, we present the multimodal procedural planning (MPP) task, in which models are given a high-level goal and generate plans of paired text-image steps, providing more complementary and informative guidance than unimodal plans.
The key challenges of MPP are to ensure the informativeness, temporal coherence, and accuracy of plans across modalities.
To tackle this, we propose Text-Image Prompting (TIP), a dual-modality prompting method that jointly leverages zero-shot reasoning ability in large language models (LLMs) and compelling text-to-image generation ability from diffusion-based models. 
\ours~ improves the interaction in the dual modalities using \tib~ and \itb~, allowing LLMs to guide the textual-grounded image plan generation and leveraging the descriptions of image plans to ground the textual plan reversely.
To address the lack of relevant datasets, we collect \wiki~ and \recipe~ as a testbed for MPP.
Our results show compelling human preferences and automatic scores against unimodal and multimodal baselines on \wiki~ and \recipe~ in terms of informativeness, temporal coherence, and plan accuracy.\footnote{Our code and data: \href{https://github.com/YujieLu10/MPP}{https://github.com/YujieLu10/MPP}}
\end{abstract}

\section{Introduction}
Recent advances in embodied~\citep{huang2022language, anderson2018vision} and conversational~\cite{qiu2021socaog} agents achieve prominent performance in task completion as humans by following instructions informed by texts and images.
However, to what extent the models can provide useful guidelines for humans to complete the task remains underexplored.
To uncover this, we propose the multimodal procedural planning task (as shown Figure~\ref{fig:teaser}). The task aims to generate goal-conditioned (e.g. ``How to make traditional szechuan chicken'') text (e.g. ``a teaspoon of light soy sauce'' explain how to marinade chicken in Step 2) and image (e.g. help identify the ingredients ``chicken, garlic, ginger, and light soy sauce'' in Step 1) plans as useful guidelines to assist humans in task completion.

Previous work~\citep{huang2022language} has explored the generation of procedural plans in text-only form. In contrast, we generate both text and image plans, which provide guidance for the agent to perform tasks that acquire complementary information from multimodal contexts.
Generating plans in both text and image form poses new challenges since the generated plans should: a) be \textit{informative} enough in both the text and image modalities, b) obey commonsense temporal \textit{coherence}, such as the order of steps, and c) achieve high plan \textit{accuracy}, indicating the complementary and alignment among multimodal plans.

Despite significant progress~\citep{zsCoT,song2022llm} in the development of large language models (LLMs), they are unable to generate images.
Existing text-to-image (T2I) models can generate high-quality images conditioned on textual instructions~\citep{ramesh2022hierarchical, rombach2022high, brooks2022instructpix2pix}.
However, they are limited in their ability to generate images that require complex text comprehension, such as temporal reasoning (e.g. ``learn basic surf safety \textit{before} hitting the waves'') and physical reasoning (e.g. ``\textit{pick up} the wine glass'').
Additionally, generating text and image plans separately using LLMs and T2I models results in inconsistency and incoherence  between the two modalities.

In this paper, we propose \method~ (\ours~), a novel dual-modality prompting framework that jointly leverages the capabilities of LLMs and T2I models for multimodal procedural planning.
We first generate vanilla text plans by directly asking LLMs~\citep{zsCoT} for step-by-step procedures.
To generate textual-grounded image plans, we devise the \tib~ (\tti~), which elicits the complex language comprehension abilities of LLMs to assist T2I models in generating informative image plans conditioned on text plans.
Similarly, we generate visual-grounded text plans using the \itb~ (\tti~), which verbalizes the image plans and injects them back into LLMs to aid in revising the text plans, thereby improving their informativeness. The temporal coherence of the generated plans is also improved  considering the context of both text and image.
Benefiting from our dual-modality prompting, our generated plans are complementary and aligned across text and image modalities.

To address the lack of suitable datasets for evaluating multimodal procedural planning, we collect the \wiki~ and \recipe~ datasets for benchmarking the task. We empirically evaluate the effectiveness of \ours~ on \wiki~ and \recipe~ in a zero-shot setting and compare it with various baselines. Our results demonstrate that \ours~ generate plausible multimodal plans that are informative, temporally coherent, and accurate.
Our work highlights the potential of combining knowledge from LLMs and T2I models to uncover multimodal zero-shot planning capabilities.
Our main contributions are as follows:
\begin{itemize}[noitemsep, topsep=0.5pt]
    \item We introduce the multimodal procedural planning task and evaluate model performance using our collected \wiki~ and \recipe~ datasets.
    \item  We propose \method~ (\ours~), a dual-modality prompting approach that elicits procedural knowledge jointly from LLMs and T2I models, enabling visual-grounded text plans and textual-grounded image plans.
    \item We show  that \ours~ substantially improves performance in terms of textual and visual informativeness, temporal coherence, and plan accuracy on human and automatic evaluations.
\end{itemize}

\begin{figure*}[t!]
    \centering
    \vspace{-3mm}
    \includegraphics[width=\textwidth]{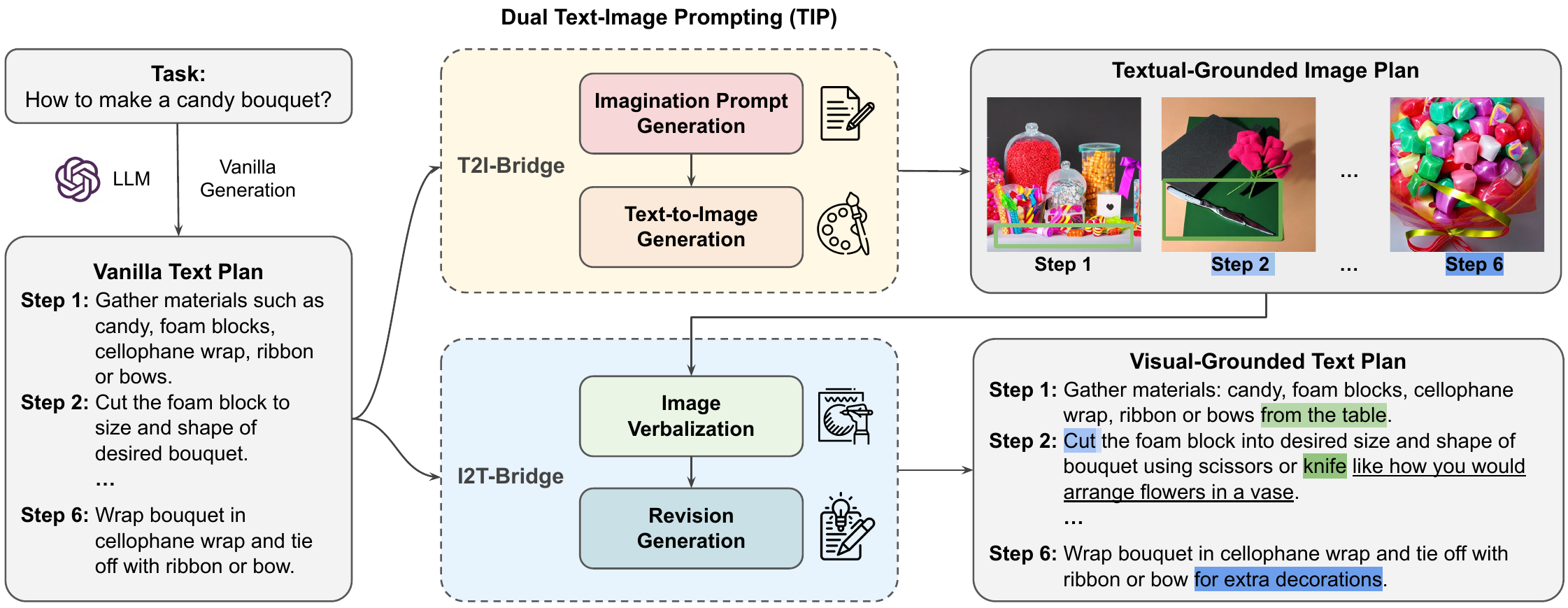}
    \caption{
    The vanilla text plan is generated using LLM. Our \method~ (\ours~) generates the textual-grounded image plan using T2I-Bridge (Fig.~\ref{fig:module_t2i_bridge}) and the visual-grounded text plan using I2T-Bridge (Fig.~\ref{fig:module_i2t_bridge}). The colors \blue{blue} and \green{green} highlight the improved grounding in text and image respectively.
    \vspace{-3mm}
    }
    \label{fig:overview}
\end{figure*}

\section{Related work}
\noindent
\paragraph{Procedural Planning} This task~\citep{zhang2020reasoning,chang2020procedure} has gain much attention in various aspects, including robotics~\cite{tellex2011understanding,jansen2020visually,brohan2022can}, vision-and-language navigation~\cite{anderson2018vision}, conversational assistants ~\cite{ilievski2018goal,qiu2021socaog,qiu2022towards,yang2022interpretable}, and animation~\cite{zhao2022triangular}. Recent work is extended to the multimodal scenarios ~\citep{wu2022understanding,song2022llm,wang2022multimedia}. In this work, we explore the multimodal procedural planning that generates goal-conditioned text and image sequences grounded in a multimodal context.

\noindent\paragraph{Multimodal Generative Models}
Recently advanced diffusion models~\cite{ramesh2022hierarchical, rombach2022high} have shown remarkable abilities in generating high-quality images given text prompts. However, generating images with desired semantics requires proper prompts, which often come from a number of trials and errors \cite{liu2022design}. To get more controllable generations, researchers have used large language models (LLMs) to expand input prompts with rich contextual knowledge. InstructPix2Pix~\cite{brooks2022instructpix2pix} combines the knowledge of GPT-3 and Stable Diffusion to generate large-scale examples of image editing as training data.
In turn, recent advances in large-scale models based on transformers~\citep{li2022blip, wang2022ofa} exhibit incredible ability in image captioning, describing the given image using natural language.

\noindent\paragraph{Injecting Visual Knowledge in LLMs}
Incorporating visual knowledge into large language models through visual imagination is a promising area of research. This can be achieved through the use of existing images as augmented visual features for language models, or through the generation of images to provide additional visual supervision to language models~\citep{yang2022z}.
Studies such as~\cite{zhang2021semi, yang2022z, zhu2022visualize, lu-etal-2022-imagination, liu2022things} have demonstrated the effectiveness of this approach.
Our proposed \ours~ exploits the image descriptions in language form to inject the visual knowledge into LLMs and elicit its potential zero-shot reasoning ability to ground the textual sentences in the verbalized visual context.

\section{Our Approach}
\subsection{Problem Definition}
\label{sec:definition}
We formulate multimodal procedural planning as a conditional text and image sequence generation problem.
Given a high-level goal $\mathcal{G}$ in natural language form, the model generates a sequence of low-level steps $\mathcal{S} = \{s_1, s_2, ..., s_n\}$.
Each step $s_i$ in the sequence is represented by a paired text $t_i$ and image $v_i$ at timestep $i$.
The text plan $\{t_1, t_2, ..., t_n\}$ and image plan $\{v_1, v_2, ..., v_n\}$ are both intended to be informative in their respective modalities and complementary across modalities.
The final multimodal procedural plans ($\mathcal{S}$) is the combination of the text plan and image plans, which describe the procedure of completing the high-level goal.

\subsection{Method Overview}
\label{sec:overview}
\begin{figure}[t]
    \centering
    \includegraphics[width=\linewidth]{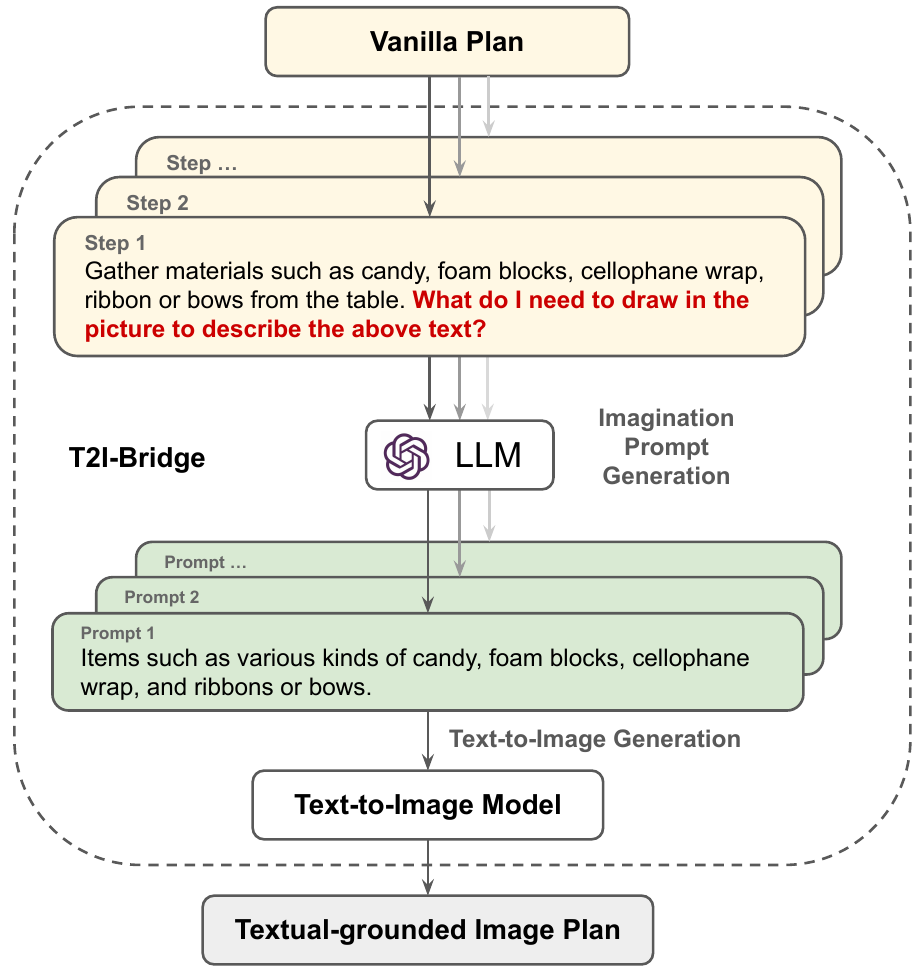}
    \caption{\tti~ elicits visual imagination in LLM to generate explicit scene description (imagination prompt) for text-to-image model conditioned on the vanilla plan.
    }
    \label{fig:module_t2i_bridge}
\end{figure}

We first elicit the zero-shot step-by-step reasoning ability in large language models (LLMs) to generate a vanilla text-only plan (left part in Figure~\ref{fig:overview}).
To enable grounding in multimodal context, we propose \method~ (\ours~), a dual-modality prompting method (middle part in \autoref{fig:overview}) upon LLMs and multimodal generative models: (1) \tib~ (\tti~): we generate the visual imaginative prompt that translates the complex textual instructions (vanilla plan in Figure~\ref{fig:module_t2i_bridge}) into explicit scene descriptions (prompt in Figure~\ref{fig:module_t2i_bridge}) for text-to-image models. (2) \itb~ (\itt~): we verbalize the image plan with the image captioning model for generating prompts (red highlighted template in Figure~\ref{fig:module_i2t_bridge}) that elicit the revision ability of LLMs with awareness of context.
Figure~\ref{fig:overview} depicts how \ours~ implements multimodal procedural planning by connecting LLMs and multimodal generative models (Image Caption Model, Text-to-Image Model) with our \tti~ and \itt~, grounding the image plansin textual context and the text plan in visual context respectively (right part in \autoref{fig:overview}).

\subsection{Vanilla Text Plan Generation}
We first elicit procedural knowledge of LLM to generate vanilla text plan using Zero-shot Chain-of-Thought~\citep{zsCoT} that does not require heavy human-engineered few-shot examples.
Specifically, we leverage InstructGPT~\citep{Ouyang2022TrainingLM} to generate a goal-conditioned step-by-step procedure with the template ``\textit{[TEMPLATE]} Task: \textit{[GOAL]}?''.
\textit{[TEMPLATE]} represents the hand-crafted template to extract the procedural knowledge from LLM. We extend the template ``Let's think step by step'' (proposed in ~\citep{zsCoT}) as ``What's the step-by-step procedure of'' for procedural planning. Then we replace the input slot \textit{[GOAL]} with the given task name $T$ (the high-level goal description) as the prompt $P$ to be fed into the LLM. The LLM then outputs goal-conditioned subsequent steps $\mathcal{W} = \{t_1, t_2, ..., t_n\}$ using greedy decoding as our initial textual plan, which is conditioned only on the task name $T$ in zero-shot generation manner.

\subsection{Textual-Grounded Image Plan Generation with \tib~}
\label{sec:t2ib}

Our Text-to-Image Bridge (\tti~) in Figure~\ref{fig:module_t2i_bridge} leverages LLM  to bridge the gap between the language understanding capabilities of LLM and the ability of language-conditioned image generation in the text-to-image model.

\noindent \textbf{Imagination Prompt Generation}
We encourage LLM to revise the prompt that already processes the physical or temporal meaning residing in the original textual plan.
To access this, for each step, we use the prompt $P_{t2i}$ ``\textit{[STEP]} \textit{[T2I-B]}'' that concatenates the original generated textual plan at step $i$ and the \tib~ template. \textit{[STEP]} represents one of the subsequent steps generated from LLMs. For \textit{[T2I-B]}, we use the trigger sentence similar to ``What do I need to draw in the picture to describe the above text?''. With this \tib~ guided prompt $P_{t2i}$, the text-to-image model then generates the textual grounded image at each timestep to compose the final sequence of visual plan $\mathcal{V} = \{v_1, v_2, ..., v_n\}$.

\begin{figure}[t]
\minipage{0.24\linewidth}
  \includegraphics[width=\linewidth]{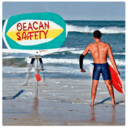}
  \subcaption{w. \tti~}
\endminipage\hfill
\minipage{0.24\linewidth}
  \includegraphics[width=\linewidth]{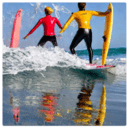}
  \subcaption{w.o. \tti~}
\endminipage\hfill
\minipage{0.24\linewidth}
  \includegraphics[width=\linewidth]{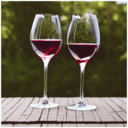}
  \subcaption{w. \tti~}
\endminipage\hfill
\minipage{0.24\linewidth}
  \includegraphics[width=\linewidth]{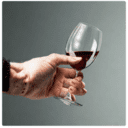}
  \subcaption{w.o. \tti~}
\endminipage\hfill
\caption{Text-to-image generation showcases with (a) (b) on ``before hitting the waves, read up on ocean safety tips and know the rules of the beach'' and (c) (d) on ``put down the wine glass'' with or without \tti~.
}
\label{fig:t2i_showcase}
\end{figure}

\noindent \textbf{Text-to-Image Generation}
We exploit the Stable Diffusion~\citep{rombach2022high} model to generate RGB images at $512 \times 512$ resolution.
Figure~\ref{fig:t2i_showcase} provides  examples of text-to-image generation with and without our \tti~.
Benefiting from the existing knowledge in LLMs, the text-to-image models are able to generate semantically relevant and high-fidelity images based on the already processed prompt.

\begin{figure}[t]
    \centering
\includegraphics[width=\linewidth]{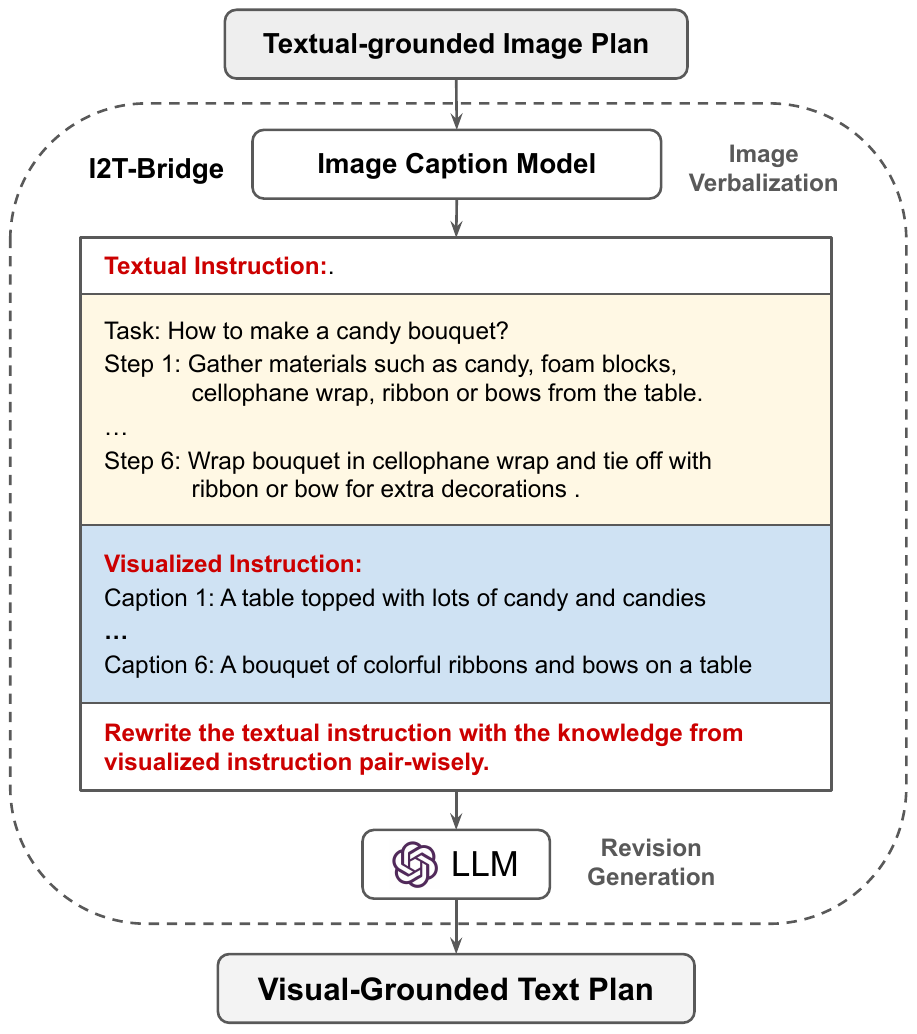}
    \caption{
    \itt~ injects verbalization of the image plans to foster revision generation of visual-grounded text plans with awareness of multimodal context.
    }
    \vspace{-5mm}
    \label{fig:module_i2t_bridge}
\end{figure}

\subsection{Visual-Grounded Text Plan Generation with \itb~}
\label{sec:i2tb}
To enhance the completeness, alignment, and knowledge exchange between the generated text and image plans, we propose revising the initial text plan based on the textual-grounded image plan.

\noindent \textbf{Image Verbalization}
To complete this, we first need to transfer the visual plan into a natural language format and then inject it into LLM.
We implement this by generating captions for each visual plan.
Given the image $v$, the captioning model BLIP~\citep{li2022blip} generates captions, which transfer the visual knowledge into textual descriptions.
For each generated visual plan $v_i$ at each timestep $i$, we generate such pairwise caption with $caption = G(v, Desc)$, where $Desc$ denotes the task description for unified vision and language models, ``what does the image describe'' in our case.
With these image captions, we can further transfer the visual-grounded information into LLM and revise our textual plan.

\noindent \textbf{Revision Generation}
To ground the textual plan in visual context, we use the verbalized description of the visual plan to concatenate with our \itb~ template similar to ``Let's revise the procedure using the captions''.
Concretely, we concatenate the initial textual plan, the captions of the visual plan, and the \itb~ template as the prompt $P_{i2t}$ ``Step-by-step Procedure: \textit{[INITIAL]} Captions: \textit{[CAPTION]} \textit{[I2T-B]}''.
In this way, we elicit the zero-shot multimodal reasoning ability of LLMs to ground the textual plan in verbalized visual context, as depicted in Figure~\ref{fig:module_i2t_bridge}
To this end, our generated multimodal plan is bi-directional grounded by connecting the abilities of LLMs and multimodal generative models.

\begin{table*}[t]
\centering
\resizebox{.95\textwidth}{!}{%
\begin{tabular}{l l ccc ccc ccc ccc}
\toprule
\multirow{2}{*}{ \textbf{Dataset}} & \multirow{2}{*}{ \textbf{Ours vs. Model}} & \multicolumn{3}{c}{\textbf{Textual-Informativeness}} & \multicolumn{3}{c}{\textbf{Visual-Informativeness}}  & \multicolumn{3}{c}{\textbf{Temporal Coherence}} & \multicolumn{3}{c}{\textbf{Plan Accuracy}} \\
\cmidrule(lr){3-5}\cmidrule(lr){6-8}\cmidrule(lr){9-11}\cmidrule(lr){12-14}
 &   & Win($\uparrow$) & Tie  & Lose($\downarrow$) & Win($\uparrow$) & Tie  & Lose($\downarrow$) & Win($\uparrow$) & Tie  & Lose($\downarrow$) & Win($\uparrow$) & Tie  & Lose($\downarrow$) \\
    \midrule
    \multirow{6}{*}{\textbf{\wiki~}} & \vgt~ & \textbf{63.34} & 18.38 & 18.27 & \textbf{60.63} & 20.45 & 18.92 & \textbf{61.95} & 21.03 & 17.02 & \textbf{61.99} & 19.40 & 18.61 \\
    &\vgtblip~ & \textbf{62.70} & 18.70 & 18.60 & \textbf{61.26} & 21.18 & 17.56 & \textbf{62.22} & 20.78 & 17.00 & \textbf{62.29} & 18.28 & 19.43 \\
    &\tgtdalle~ & \textbf{62.61} & 20.34 & 17.06 & \textbf{59.88} & 22.38 & 17.74 & \textbf{60.53} & 22.08 & 17.40 & \textbf{61.19} & 22.07 & 16.74 \\
    &\tgt~  & \textbf{62.58} & 19.82 & 17.60 & \textbf{60.25} & 21.16 & 18.58 & \textbf{60.68} & 22.38 & 16.94 & \textbf{61.73} & 20.56 & 17.72 \\
    &\utwo~  & \textbf{60.68} & 21.56 & 17.76 & \textbf{59.90} & 20.41 & 19.70 & \textbf{60.22} & 22.99 & 16.79 & \textbf{60.41} & 21.53 & 18.06 \\
    &\uthree~ & \textbf{62.32} & 19.82 & 17.86 & \textbf{60.29} & 20.85 & 18.85 & \textbf{61.10} & 22.17 & 16.73 & \textbf{61.48} & 20.29 & 18.23 \\

    \midrule
    \multirow{6}{*}{\textbf{\recipe~}} & \vgt~ & \textbf{64.51} & 18.29 & 17.20 & \textbf{62.39} & 20.18 & 17.43 & \textbf{62.74} & 20.40 & 16.86 & \textbf{63.66} & 19.19 & 17.15 \\
    &\vgtblip~ & \textbf{64.81} & 18.58 & 16.61 & \textbf{62.29} & 19.60 & 18.11 & \textbf{62.70} & 20.72 & 16.58 & \textbf{62.90} & 19.08 & 18.02 \\
    &\tgtdalle~ & \textbf{61.16} & 20.15 & 18.69 & \textbf{59.60} & 20.60 & 19.80 & \textbf{60.04} & 20.48 & 19.48 & \textbf{62.11} & 19.21 & 18.68 \\
    &\tgt~ & \textbf{61.31} & 19.81 & 18.87 & \textbf{60.49} & 20.37 & 19.14 & \textbf{60.37} & 20.33 & 19.31 & \textbf{62.38} & 18.81 & 18.81 \\
    &\utwo~ & \textbf{62.50} & 19.33 & 18.17 & \textbf{60.59} & 18.12 & 21.29 & \textbf{61.24} & 21.13 & 17.63 & \textbf{62.30} & 17.38 & 20.31 \\
    &\uthree~ & \textbf{62.65} & 19.26 & 18.09 & \textbf{61.10} & 20.00 & 18.90 & \textbf{61.46} & 20.60 & 17.94 & \textbf{62.85} & 18.75 & 18.40 \\

    \bottomrule
\end{tabular}
}
    \caption{Percentages of multimodal procedural planning results of \ours~ that are better than, tied with, or worse than baselines, on randomly sampled $200$ distinct tasks from each dataset.} 
    \label{tab:human_winlose}
\end{table*}

\section{Experiments}
\subsection{Datasets}
Our datasets are collected and repurposed from \wikihow~\footnote{\url{https://www.wikihow.com}} and \recipeqa ~\citep{yagcioglu2018recipeqa} due to their temporal relatedness among texts and images.
We collect \wiki~ by crawling the household ``how to'' articles from \wikihow~ and then repurpose them into a multimodal procedural planning dataset by formulating the article title as the task name and content as the textual steps, with the pictures as the visual steps.
\recipeqa~ is a dataset designed for multimodal comprehension of cooking recipes.
We collect \recipe~ from this dataset for multimodal procedural planning by sequencing all the given text-image pairs as the text and image plan correspondingly, with the main title as the task name.
We conduct zero-shot experiments on $1,000$ distinct, randomly sampled tasks from each dataset.
Please refer to Appendix~\ref{app:dataset_details} for more details on the datasets.

\begin{table*}[!htb]
\centering
\resizebox{\textwidth}{!}{%
\begin{tabular}{l l ccccccccccc}
\toprule
\multirow{3}{*}{\textbf{Dataset}}& \multirow{3}{*}{\textbf{Model}}& \multicolumn{4}{c}{\textbf{Text Plan}} & \multicolumn{2}{c}{\textbf{Image Plan}} & \multicolumn{3}{c}{\textbf{Multimodality Plan}} & \multicolumn{1}{c}{\textbf{Step Length}}\\
\cmidrule(lr){3-6}\cmidrule(lr){7-8}\cmidrule(lr){9-11}\cmidrule(lr){12-12}\
 & & WMD & S-BERT & ROUGE-L & METEOR  & FID $\downarrow$ & CLIP $\uparrow$  & Cap-S & Text-S & ALL-S & Avg. \\
    \midrule
    \multirow{7}{*}{\textbf{\wiki~}}  & \vgtblip~         &0.78&0.35&0.06&0.04 & -&0.71 &\underline{0.36}&0.41&\underline{0.39}& 8.26  \\
    & \vgt~        &0.86&0.27&0.07&0.06 & - &0.71 &0.27&\underline{0.48}&0.37& 8.26 \\
    \cmidrule(lr){2-12}\
    & \tgtdalle~                     &0.68&\underline{0.76}&\underline{0.28}& \underline{0.12}& \underline{47.39} &0.74&0.33&0.26&0.29& 8.26 \\
    & \tgt~                           &0.68&0.76&0.28& 0.12& 56.64 &0.73 &0.34&0.26&0.30& 8.26 \\
    \cmidrule(lr){2-12}\
    & \utwo~                                    &\underline{0.87}&0.65&0.10&0.06 &61.17&0.50 &0.33&0.25&0.28& 4.70  \\
    & \uthree~                                &0.86&0.67&0.11&0.08 &57.87 &0.70 &0.33&0.27&0.30& 6.68 \\
    \cmidrule(lr){2-12}\
    & \textbf{\mplan}~                                &\textbf{0.90} &\textbf{0.67} & \textbf{0.12} &\textbf{0.09} &\textbf{48.82} &\underline{\textbf{0.78}} & \textbf{0.34}&\textbf{ 0.28} & \textbf{0.31} & 6.75  \\
    \midrule
    \multirow{7}{*}{\textbf{\recipe~}} & \vgtblip~         &0.77&0.37&0.08&0.05 &- &0.64&0.42&\underline{0.56}&\underline{0.49}&6.93   \\
    & \vgt~                      &0.82&0.40&0.09 &0.10 & -& 0.64&0.43&0.48&0.46&6.93   \\
    \cmidrule(lr){2-12}\
    & \tgtdalle~                 &0.21 &0.59 &0.10 &0.09 &\underline{53.55} &0.63&0.46&0.40&0.43&6.93   \\
    & \tgt~                      &0.21 &0.59 &0.10 &0.09&54.58 &0.61 &\underline{0.48}&0.40&0.44& 6.93  \\
    \cmidrule(lr){2-12}\
    & \utwo~                     &0.84&0.63&0.11&0.10 &60.11 &0.49 &0.44&0.33& 0.38 &5.17  \\
    & \uthree~                   &0.85&0.68&0.12&0.13 &60.07 &0.73&0.42&0.35&0.38&6.82  \\
    \cmidrule(lr){2-12}\
    & \textbf{\mplan}~                    &\underline{\textbf{0.86}} & \underline{\textbf{0.68}}& \underline{\textbf{0.13}} &\underline{\textbf{0.14}} & \textbf{58.68}&\underline{\textbf{0.79}}&\textbf{0.43}&\underline{\textbf{0.36}}&\underline{\textbf{0.40}}& 6.94  \\
    \bottomrule
\end{tabular}
}
    \caption{
    Automatic evaluations on $2,000$ distinct tasks from \wiki~ and \recipe~. Image Ref and Text Ref baselines use image and text title references from the dataset. Our \ours~ uses \dthree~ and Stable-Diffusion as the LLM and T2I model. We \underline{underline} and \textbf{bold} highest score of models with and without reference baselines.
} 
    \label{tab:auto_eval}
\end{table*}

\subsection{Evaluation Metrics}
We conduct head-to-head comparisons using Amazon Mechanical Turk (AMT) platform (details can be found in Appendix~\ref{app:amazon_mechanical_turk}) on four aspects: (1) \texttt{Textual Informativenss}: the text plans contain the necessary information to complete the task, (2) \texttt{Visual Informativeness}: the image plans contain the necessary information to complete the task, (3) \texttt{Temporal Coherence}: the multimodal plans meet the temporal commonsense requirements, such as the order in which the steps occur, (4) \texttt{Planning Accuracy}: whether referring to the multimodal plans can successfully assist task completion.
In addition, we measure semantic relevance between predicted text plans and reference text plans using Word Mover's Distance (WMD)~\citep{pmlr-v37-kusnerb15}, Sentence-BERT (S-BERT)~\citep{Reimers2019SentenceBERTSE}, ROUGE-L~\citep{lin-2004-rouge}, and METEOR~\citep{Banerjee2005METEORAA}.
We measure FID score~\citep{FID} and CLIP score~\citep{hessel-etal-2021-clipscore, Radford2021LearningTVCLIP} of image plans.
We compute S-BERT between the captions of the predicted image plan and the reference text plan as the Caption-Sentence-BERT score (Cap-S), and between the predicted text plan and the reference text plan as the Text-Sentence-BERT score (Text-S). We then average these two scores to obtain the All-Sentence-BERT score (ALL-S) for multimodal plans.
Evaluations are conducted at a procedure level.

\begin{table*}[!ht]
\centering
\resizebox{\textwidth}{!}{%
\begin{tabular}{l cc | l cc}
\toprule
\multirow{2}{*}{\textbf{\tib~ Template}} & \multicolumn{2}{c}{\textbf{Alignment}} & \multirow{2}{*}{\textbf{\itb~ Template}} & \multicolumn{2}{c}{\textbf{Alignment}} \\
\cmidrule(lr){2-3}\cmidrule(lr){5-6}\
& \wiki~ & \recipe~ & &\wiki~ & \recipe~ \\
\midrule
    \color{purple}{What do I need to draw in the picture to describe the above text?} & \textbf{0.9625} & \textbf{0.9595}& \color{purple}{Rewrite the textual instruction with the knowledge from visualized instruction pair-wisely.} & \underline{0.7644}  & \underline{0.6945}\\
    What do you see in the figure? & \underline{0.9366} & \underline{0.9397} & Based on the visual caption, can you revise the step-by-step procedure according to the paired captions? &\textbf{0.8011} & 0.6205\\
    Describe what the picture corresponding to the text should have. & 0.9070 & 0.9181 &  Revise each step according to the visual imagination. &0.6921 & 0.7329\\
    Let's think about what we need to visualize to present the above idea. &0.8986 & 0.8941 & Let's revise the procedure using the captions. & 0.6155& \textbf{0.7691}\\
    \midrule
    \underline{Describe something irrelevant to the above text.} &0.5598 & 0.5325&  \underline{What's the procedure that disobey the captions?} &0.5079 &0.5902\\\
    \underline{What do you usually draw?} &0.5350 &0.4826 & \underline{Provide an interesting procedure to be irrelevant with the captions.} &0.1519 & 0.163 \\
    \bottomrule
\end{tabular}
}
    \caption{Robustness check of various templates used in both \tib~ and \itb~ over \wiki~ and \recipe~ dataset. The \underline{underlined} templates are misleading examples. Our \method~ model chooses the template with averaged best multimodal alignment, highlighted in \color{purple}{purple}\color{black}.}
    \label{tab:robustness}
\end{table*}

\begin{figure*}[t!]
\minipage{\textwidth}%
  \includegraphics[width=\linewidth]{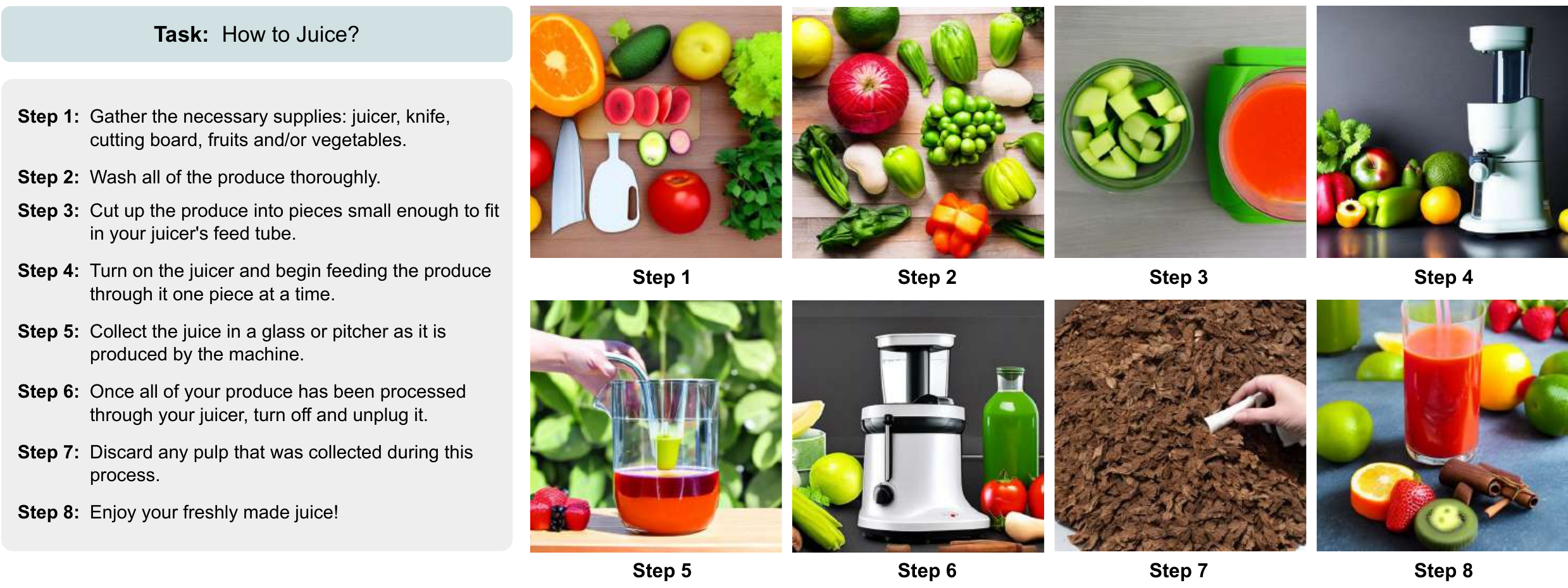}
  \subcaption{Multimodal procedural plan generated by baseline \uthree~.}
\endminipage\hfill
\vspace{1mm}
\minipage{\textwidth}%
  \includegraphics[width=\linewidth]{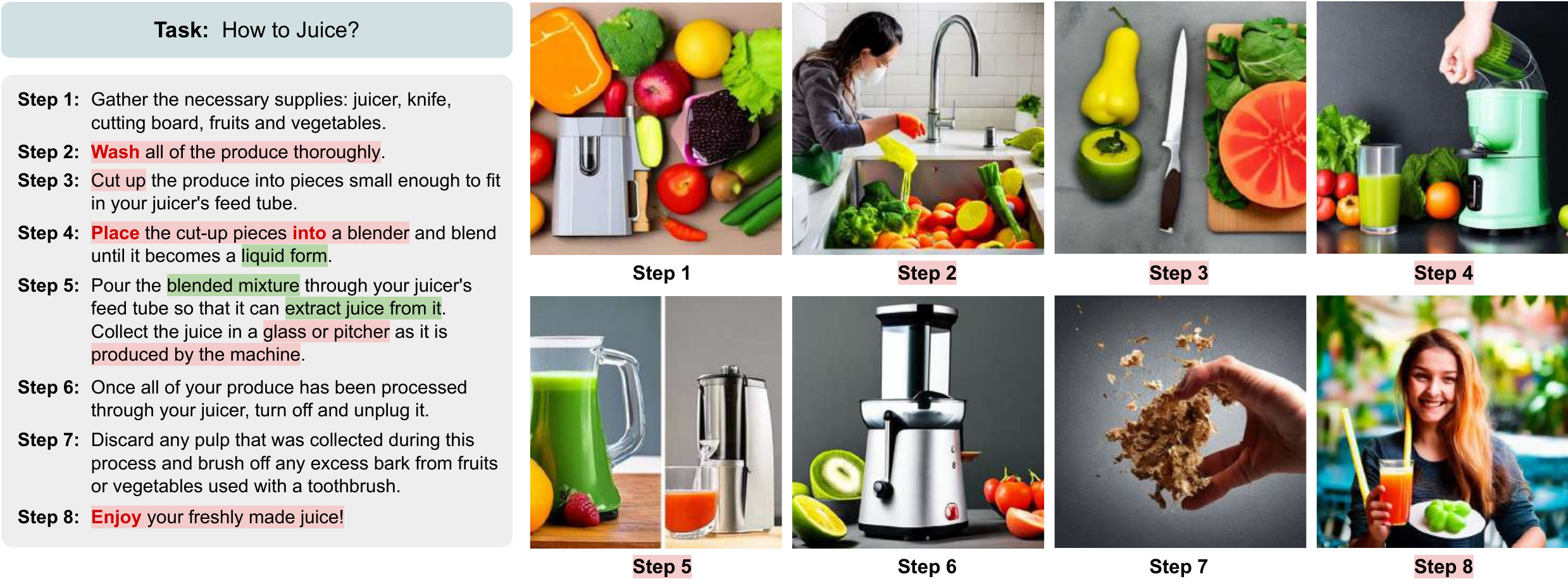}
  \subcaption{Multimodal procedural plan generated by our \method~ (\ours~).}
\endminipage\hfill
\caption{Improved grounding in textual and visual context are highlighted in \pink{pink} and \green{green} respectively. \red{Red} texts indicate reasoning of physical action in image plan generation.
}\label{fig:plan_comparison}
\end{figure*} 

\subsection{Baselines}
The key ingredient of our proposed method \ours~ is that LLMs and multimodal generative models will collaboratively generate multimodal procedural plans benefiting from our designed dual bridges: \tib~ and \itb~.
We compare \ours~ with the following baselines: (1) ImageRef + OFA/BLIP-Caption: use image plans references directly from the dataset, and generate text plans using image caption models over references (2) TextRef + DALLE/Stable-Diffusion: use title references from dataset as text plans, and use text-to-image models to generate text plans (3) Text-Davinci-002/003 + Stable-Diffusion: separately generate text and image plans using LLMs and text-to-image models (4) \step~: instead of generating the plan at the procedure level, it generates each step iteratively by prompting generated plans from the history sequences to LLMs.

\subsection{Quantitative Analysis}
\noindent \textbf{Human Evaluation Results}
We conduct Win-Tie-Lose Comparison between \ours~ and the baselines over \wiki~ and \recipe~.
Averaged results from $200$ tasks rated by $3$ crowdsourcing per example are reported in Table~\ref{tab:human_winlose}.
Across four aspects, \ours~ receives consistently higher preferences, outperforming the baselines over the winning ratio by over $60\%$.
In terms of textual informativeness, the unimodal baselines (\vgt~ and \vgtblip~) is slightly worse than the unimodal text reference based baseline (\tgt~ and \tgtdalle~) and multimodal baselines (\uthree~ and \utwo~). This is mainly due to the other baselines either direclty leverage the textual information from the reference or the rich text-based knowledge in LLMs.
In terms of visual informativeness, the multimodal baselines (\uthree~ and \utwo~) can not achieve on par results with textual reference-based baseline. We hypothesize this is due to the lack of visual knowledge injected into LLMs.
The large performance gain of \ours~ over multimodal baselines (\uthree~ and \utwo~) that simply combine the knowledge from LLMs and multimodal generative models imply the importance of grounding our multimodal plans in a multimodal context.

\noindent \textbf{Automatic Evaluation Results}
In Table~\ref{tab:auto_eval}, \ours~ achieves consistent improvement over baselines (without Ref), and even surpasses the baselines using reference from the dataset on \recipe~.
This further confirms our superiority in generating multimodal plans with semantic correctness and alignment.
Notice that Text Ref baselines directly use the title from the dataset, which is a summarized version of the main content (golden reference used in automatic evaluations).

\noindent \textbf{Template Robustness}
In Table~\ref{tab:robustness}, we compare various similar templates for \tti~ and \itt~ against misleading templates.
The Alignment is measured with CLIP~\citep{Radford2021LearningTVCLIP} to capture the similarity between given text/image and conditionally generated image/text.
The poor alignment of misleading templates and similar alignment of various bridge templates prove the robustness of the template choice in the experiments.

\subsection{Qualitative Analysis}
\noindent \textbf{Multimodal Grounding}
In Figure~\ref{fig:plan_comparison}, we compare the performance of \ours~ to baselines in multimodal procedural planning.
\ours~ generate image plans that are grounded in the textual context.
With the help of LLMs reasoning in the temporal dimension, we transfer this ability to image generation, conditioning on the revised prompts of LLMs. This allows digestion of the temporal and complex reasoning present in the text plan and directly indicates what needs to be depicted in the image.
The highlighted steps of image plans correctly visualize the scene described in the textual context. For example, at Step $2$, instead of only showing the vegetables, ours show an image of a person washing the produce thoroughly.
\ours~ also generate text plans that are better grounded in the image plan. 
The text plan correctly refers to the objects in visual input, such as ``liquid form'' and ``blended mixture'', and also complements the visual context, such as ``extract juice from it''.
Our results indicate the potential for uncovering multimodal reasoning capabilities in LLMs, even though they are primarily used for language reasoning. We provide more comparisons on multimodal procedural planning in Appendix~\ref{app:showcases}.

\subsection{Ablations}
\noindent \textbf{Bridge Effect}
We report the performance drop of \ours~ without \itb~ on average textual metrics in Table~\ref{tab:i2t_bridge}, indicating that the text plan without condition on visual information is vulnerable in text-only planning quality.
Then we ablate the text-to-image models in Table~\ref{tab:t2i_bridge}. With obvious improvement over both FID score and Alignment, we show that the \tti~ is essential to generate textual-grounded image plans.

\begin{table}[t!]
\centering
\resizebox{\columnwidth}{!}{%
\begin{tabular}{l cc}
\toprule
\multirow{2}{*}{\textbf{Plan w.o. \itt~}} & \textbf{\wiki~} & \textbf{\recipe~}\\
\cmidrule{2-2}\cmidrule{3-3}\
& Avg. Textual & Avg. Textual \\
\midrule
    Imagination Prompt & 0.341 (-18.4\%) & 0.363 (-14.1\%) \\
    Image Verbalization & 0.261 (-37.5\%) & 0.273 (-35.4\%) \\
    \bottomrule
\end{tabular}
}
    \caption{Ablation of \itb~ on \ours~ over text plan generation.
    }
    \label{tab:i2t_bridge}
\end{table}

\begin{table}[t!]
\centering
\resizebox{\columnwidth}{!}{%
\begin{tabular}{l cccc}
\toprule
\multirow{2}{*}{\textbf{Model}} & \multicolumn{2}{c}{\textbf{\wiki~}} & \multicolumn{2}{c}{\textbf{\recipe~}} \\
\cmidrule(lr){2-3}\cmidrule(lr){4-5}\
&  FID $\downarrow$ &  Align $\uparrow$ &  FID $\downarrow$ &  Align $\uparrow$ \\
\midrule
    DALLE & 119.03 & 0.77 & 83.27 & 0.64\\
    $+$ \tib~  &\textbf{117.02} & \textbf{0.79} & \textbf{67.64} & \textbf{0.78}\\
    \midrule
    Stable Diffusion & 129.13 & 0.74 & 88.17 & 0.62\\
    $+$ \tib~ & \textbf{119.74} & \textbf{0.78} & \textbf{84.37} & \textbf{0.78}\\
    \bottomrule
\end{tabular}
}
    \caption{Ablation of \tib~ with \tgt~ on single-step image plan generation.}
    \label{tab:t2i_bridge}
\end{table}

\begin{figure}[t!]
    \centering
    \includegraphics[width=\columnwidth]{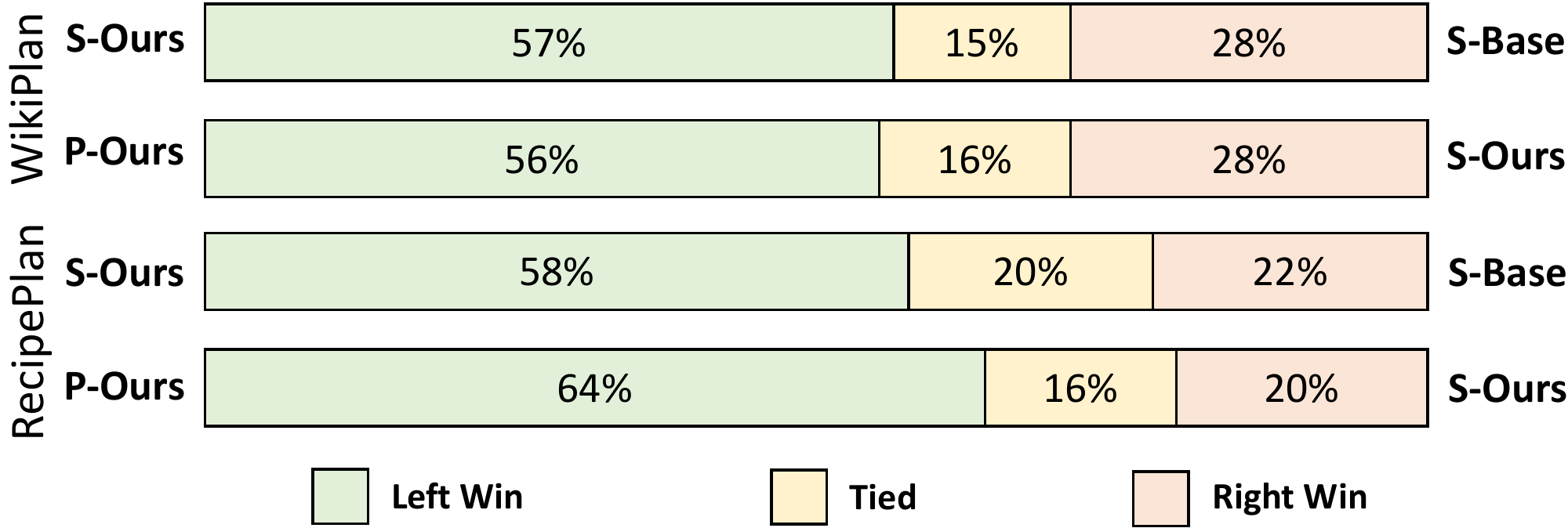}
    \caption{
    Step-based (S) vs. Procedure-based (P) Win-Tie-Lose over \texttt{Plan Accuracy}.
    }
    \label{fig:closed_loop}
\end{figure}

\noindent \textbf{Step-based or Procedure-based}
We explore our procedure-based method (P-Ours) against the step-based \ours~ (S-Ours) and step-based \uthree~ (S-Base).
We report the head-to-head comparison results on \texttt{Plan Accuracy} in Figure~\ref{fig:closed_loop}.
The procedure-based method achieves $60\%$ win rate over the step-based \ours~.
We observe this is partially due to the instinct of LLMs to repeat input texts and is less clear to understand the full intent of generation expectation. Thus the procedure-based method usually achieves better planning quality at the very beginning.
We also show that our devised \tti~ and \itt~ achieve consistent improvement on step-based mehtod with averaged winning rate of $58\%$ (S-Ours vs. S-Base).

\section{Conclusion and Future Work}
We introduce the Multimodal Procedural Planning task that aims to generate goal-conditioned text and image subsequences and benchmark models' performance with our curated testbed \wiki~ and \recipe~.
We propose Text-Image Prompt (\ours~), a dual-modality prompting framework, that connects LLMs with multimodal generative models to enable plausible multimodal procedural plan generation.
Our evaluation benchmark is limited in that no perfect metrics exist to quantify the quality of text-image plans.
We hope our work shed light on research into uncovering this limitless capability of multimodal procedural planning driven by uniform automatic metrics.

\section*{Limitations}
Relying on the LLMs to reason over complex text for text-to-image models though improving the quality, still remains a large gap with human performance. This is mainly restricted by the pre-training gap between LLMs and text-to-image models. To solve this, further work should explore the finetuning stage that how to inject this language reasoning into the multimodal generation models.

In addition to the model-side limitations, the dataset is limited in that not all the possible multimodal plans are provided and their quality is hard to validate. Due to the lack of perfect metrics in evaluating the text-image sequences, the research in multimodal procedural planning maybe difficult to scale up. Future work should explore this promising direction and furthermore lead LLMs and T2I models better multimodal procedural planners.

\section*{Ethics Statement}
We acknowledge that our research utilizes resourceful knowledge in large-scale pre-trained models, which have the potential to bias to a certain cultural background. For example, the task from \recipe~ and \wiki~ that involve food preparation may have different procedures depending on different individuals' eating habits. We encourage future studies that uncover the multimodal procedural planning ability with consideration of personalized decision makings.

The data annotation part of the project is classified as exempt by Human Subject Committee via IRB protocols.
The hourly wage paid to participants is estimated at \$12, which is higher than the federal minimum wage.
We manually ensure no personal information is collected and no offensive content is presented during human evaluations.


\bibliography{anthology,custom}
\bibliographystyle{acl_natbib}
\clearpage
\appendix
\section{Background}
A line of work in unimodal procedural planning studies sorting a series of unordered texts or events~\cite{chen2016neural,cui2018deep,oh2019topic,calizzano2021ordering,wu2022understanding}. Other work explores generating subsequent steps given a target goal, e.g.,
\citet{lu2022neuro} aim at generating a sequence of plans to complete the high-level task.

Text-to-image generation is a task that synthesizes images from text prompts.
DALL-E 2~\cite{ramesh2022hierarchical} and Stable Diffusion~\cite{rombach2022high} are state-of-the-art text-to-image models developed on top of diffusion models conditioned on input texts.
Some early work in text-to-image models trains generative adversarial networks (GANs)~\cite{goodfellow2014generative} on image captioning datasets \cite{xu2018attngan, zhu2019dm, tao2020df, zhang2021cross, ye2021improving} to generate images conditioned on textual descriptions. Other work follows the VQ-VAE~\cite{van2017neural} framework and trains autoregressive transformers that take both the text and image as sequences of tokens \cite{ramesh2021zero,ding2021cogview, gafni2022make}.
However, these methods are struggling to generate photorealistic images. Motivated by the remarkable progress of diffusion models in generating images with fidelity~\cite{sohl2015deep,song2019generative,ho2020denoising}, recent work has applied them to text-to-image generation with auxiliary text encoders~\cite{rombach2022high,nichol2021glide,gu2022vector,ramesh2022hierarchical,saharia2022photorealistic}.
\citet{wang2022diffusiondb} propose the first large-scale text-to-image prompt dataset, DiffusionDB, which enables a new research direction of prompt engineering to construct better prompts. \citet{chakrabarty2022} use GPT-3 to generate a detailed textual description with rich visual metaphors to prompt the DALL-E 2 model.

Recently, there has been a trend of using large language models (LLMs) like GPT-3~\cite{brown2020language} to transfer visual knowledge in order to improve their capabilities in downstream natural language processing (NLP) and multimodal tasks. For example, images and videos can be translated into captions which further instruct a language model to generate contextual descriptions~\cite{wang2022language, zeng2022socratic} or answer knowledge-based visual questions~\cite{yang2022empirical}. Instead of being prompted with textual descriptions, language models can extend to vision-language settings through text generation controlled by visual features~\cite{cho2021unifying, tsimpoukelli2021multimodal, su2022language, zhu2022uni, wang2022unifying, alayrac2022flamingo}.

\section{Method Details}
\subsection{Configurations}
\label{app:configs}
The experiments using Text-Davinci and DALLE are conducted with OpenAI API on January 2023. We use BLIP w/ ViT-B and CapFilt-L\footnote{\url{https://github.com/salesforce/BLIP}} and OFA-base from huggingface demo\footnote{\url{https://huggingface.co/OFA-Sys/ofa-base}}.

\subsection{Details of Module Outputs}
\label{app:inter_output}
\begin{figure*}[!htb]
    \centering
    \minipage{\textwidth}%
      \includegraphics[width=\linewidth]{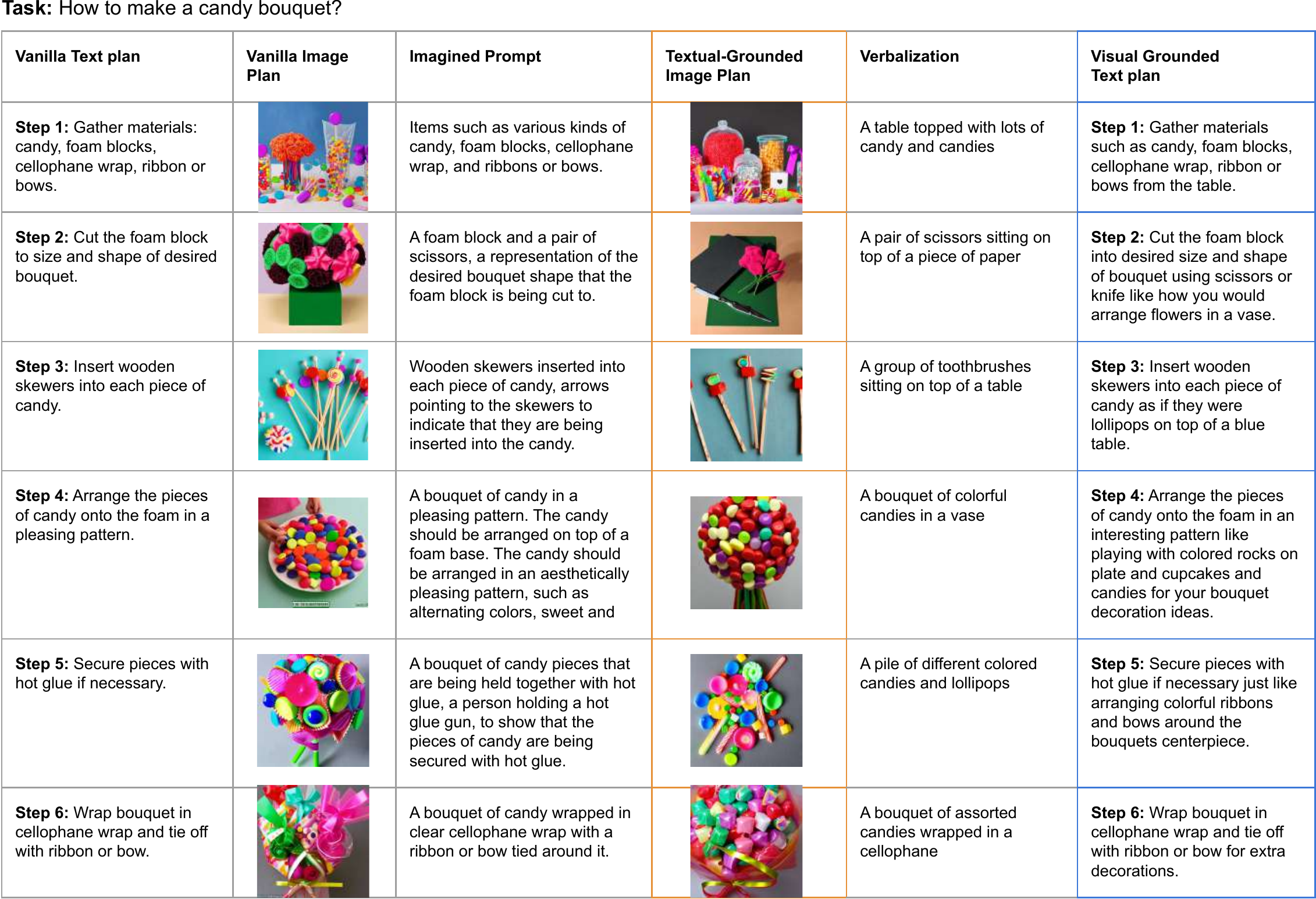}
      \subcaption{Full example outputs.}
    \endminipage\hfill
    \vspace{2mm}
    \minipage{\textwidth}%
      \includegraphics[width=\textwidth]{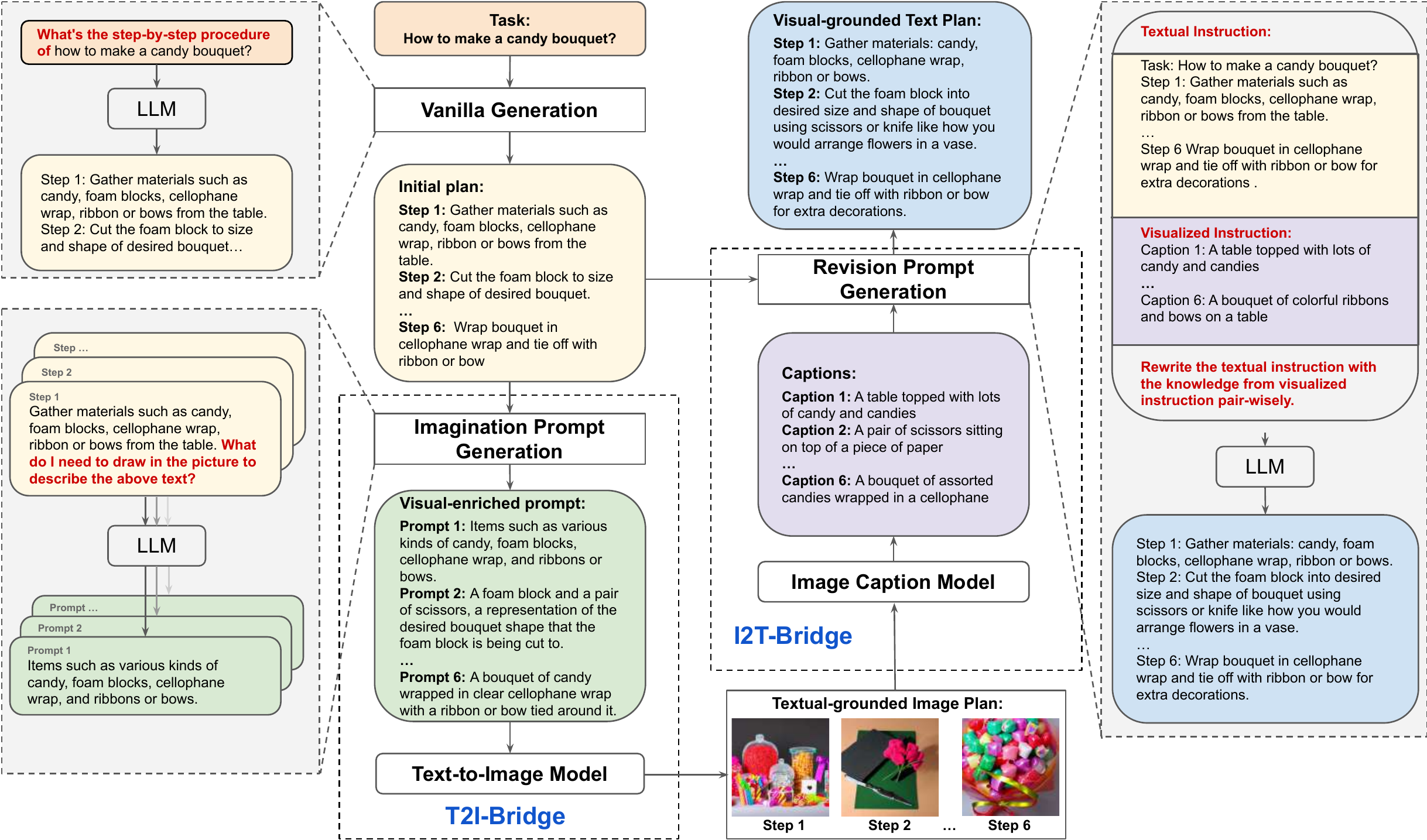}
      \subcaption{Visualization of outputs from each module.}
    \endminipage\hfill
    \caption{
    Output Details of each module of \ours~ for Multimodal Procedural Planning. 
    The \tti~ leverage the complex language comprehension and zero-shot reasoning ability of LLMs to improve text-to-image generation. Reversely, the \itt~ injects visual knowledge via verbalization of the visual plans to generate a visually-grounded and complementary textual plan.
    }
    \label{fig:output_detail}
\end{figure*}

\begin{figure*}[!htb]
    \centering
    \minipage{\textwidth}%
      \includegraphics[width=\linewidth]{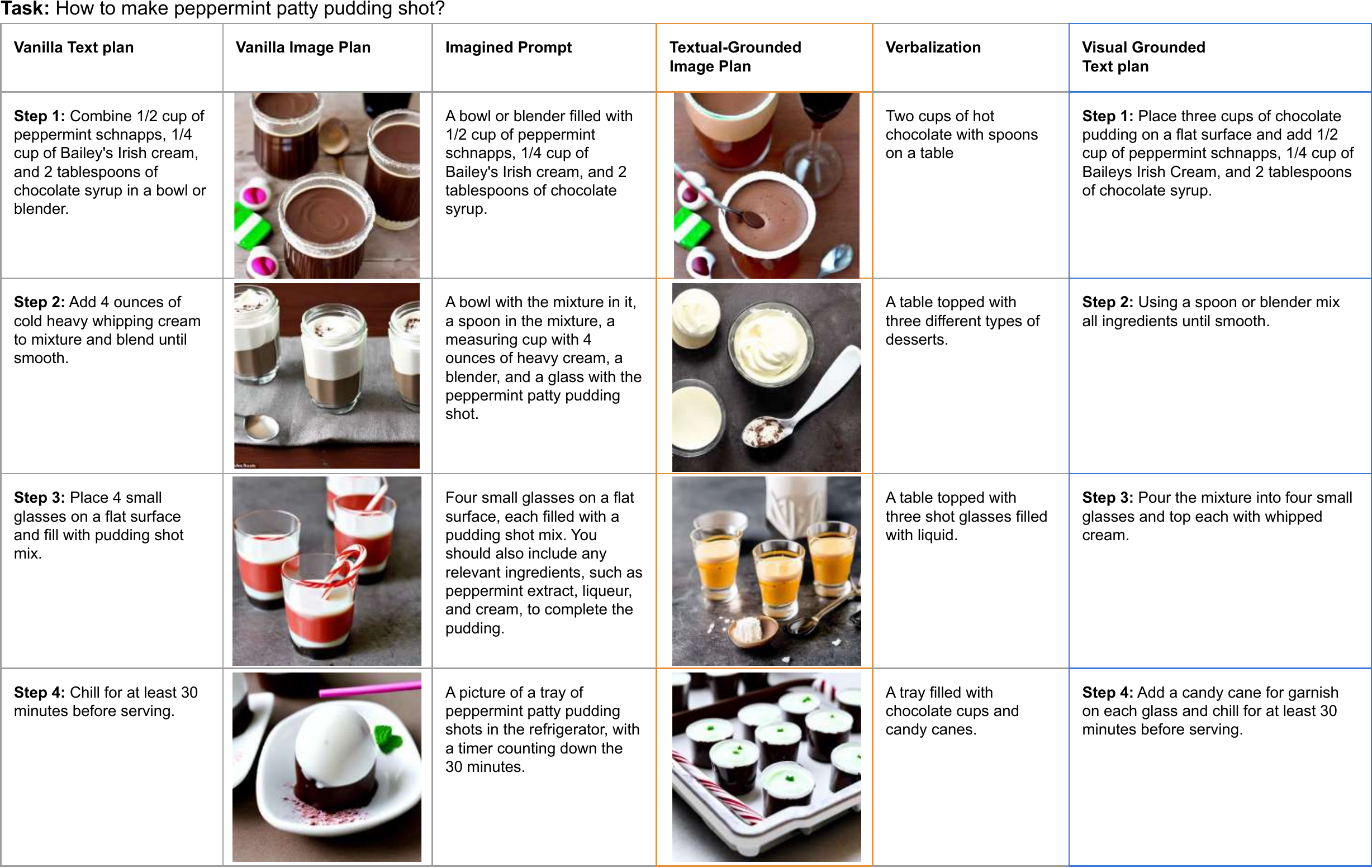}
    \endminipage\hfill
    \minipage{\textwidth}%
      \includegraphics[width=\textwidth]{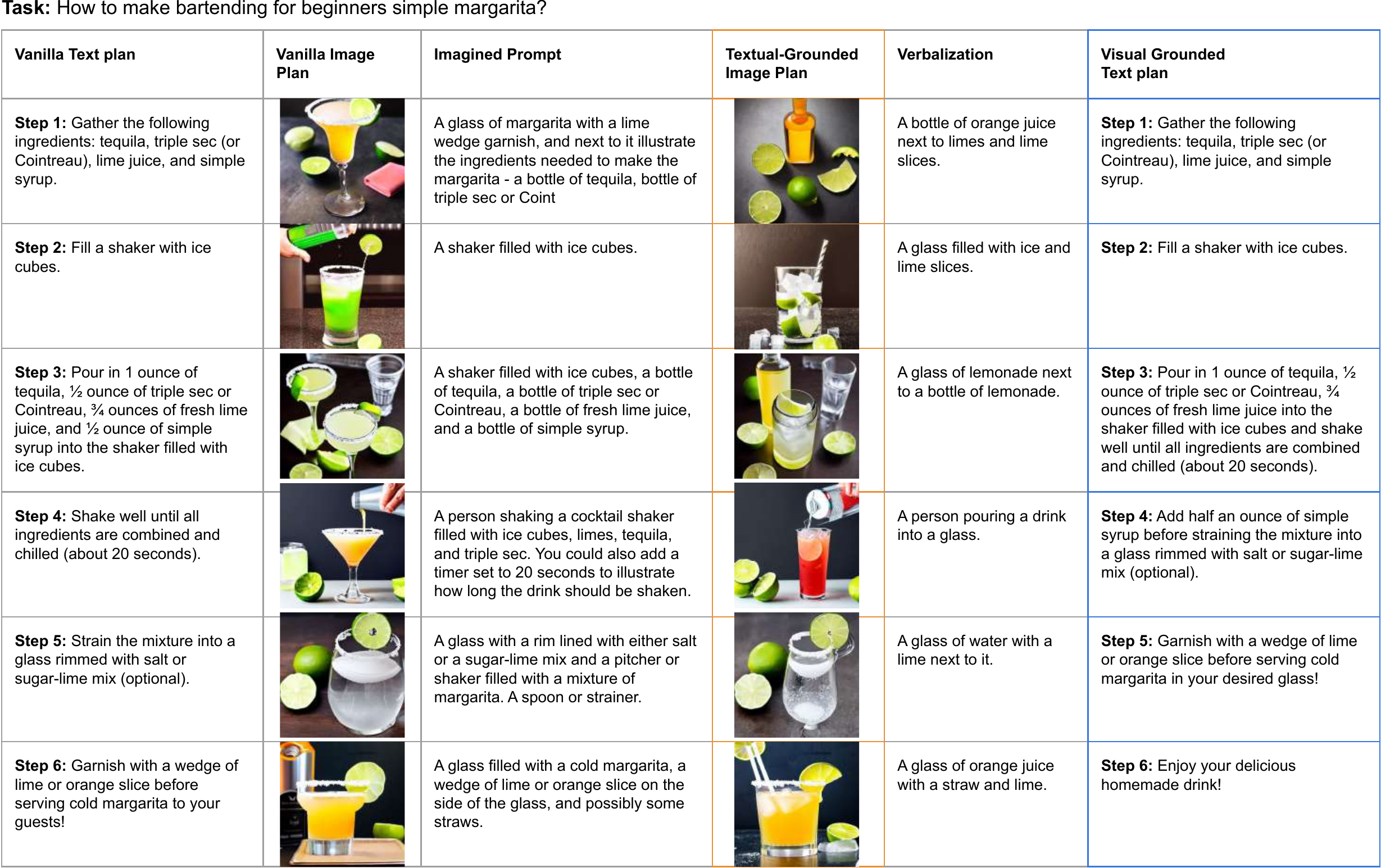}
    \endminipage\hfill
    \caption{
    Full example of outputs details of \ours~ for Multimodal Procedural Planning.
    }
    \label{fig:output_detail_more}
\end{figure*}
We showcase all the details of the outputs of each module for the example task ``How to make a candy bouquet'' in Figure~\ref{fig:output_detail}. In addition, we showcase examples on ``How to make peppermint patty pudding shot'' and ``How to make bartending for beginners simple margarita'' in Figure~\ref{fig:output_detail_more}.

\section{Dataset Details}
\label{app:dataset_details}
\subsection{\recipe~}
\noindent\textbf{Data Repurpose} 
\recipeqa~ was proposed in ~\citep{yagcioglu2018recipeqa} that provide four tasks (Textual Cloze, Visual Cloze, Visual Ordering, Visual Coherence) for multimodal machine comprehension of cooking recipes.
This dataset contains question-answer pairs generated from copyright-free recipes. Each of them is under a license, which is provided in each data JSON file.
We collect \recipe~ by repurposing the test dataset from \recipeqa~ that relates to the Visual Ordering task as the sequence generation task for the multimodal procedural planning evaluation testbed.
We use recipe instructions as textual plan reference and their paired images as visual plan reference.

\begin{figure*}[!htb]
\minipage{0.99\textwidth}
\includegraphics[width=0.99\textwidth]{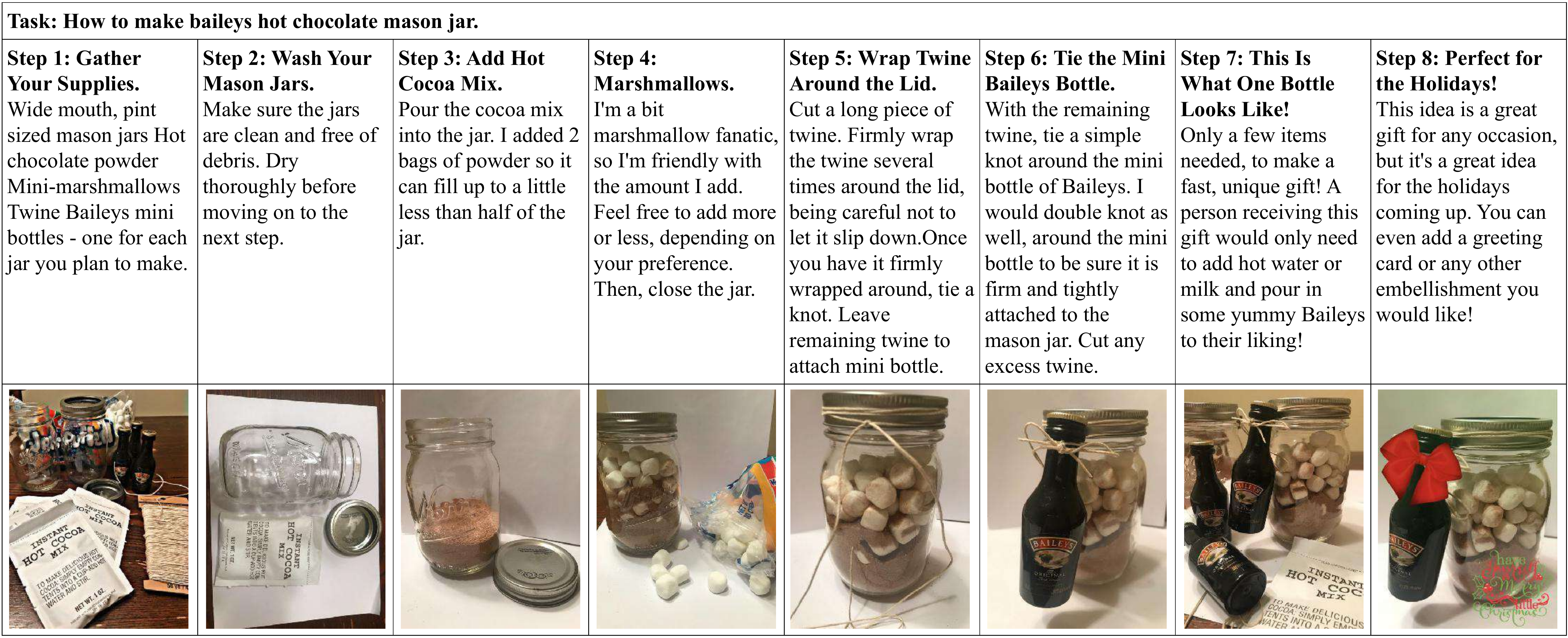}
\endminipage
\hfill
\vspace{5mm}
\minipage{0.99\textwidth}
\includegraphics[width=0.99\textwidth]{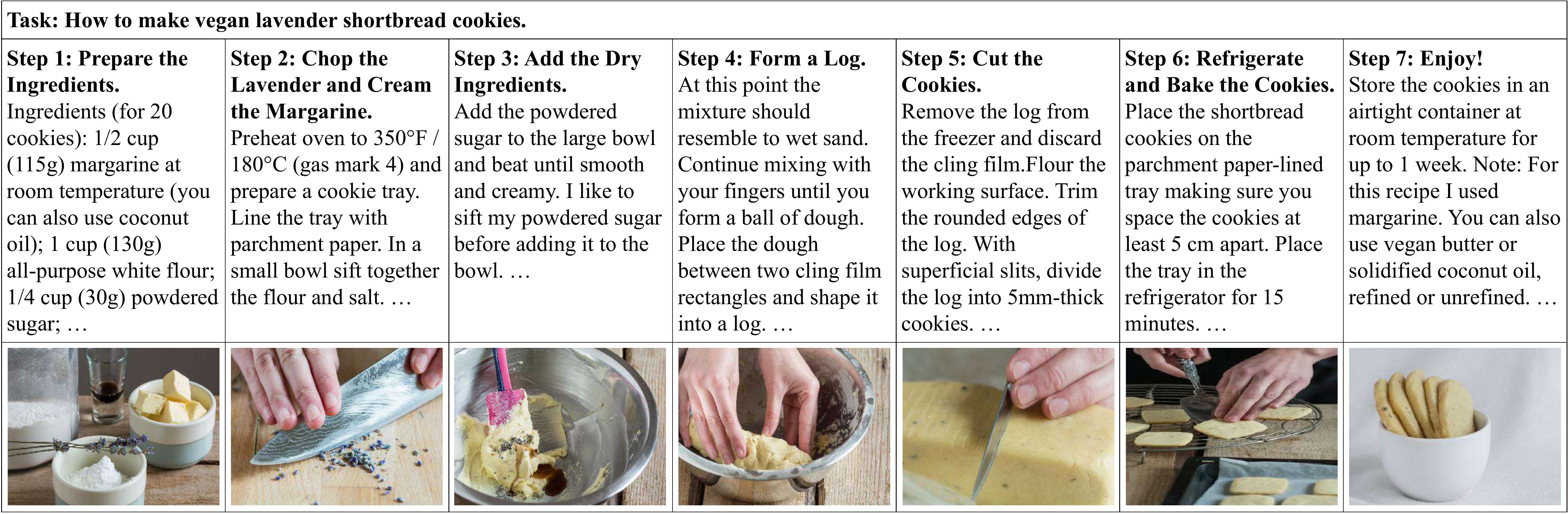}
\endminipage
\caption{Two examples in the \recipe~ dataset.}
\label{fig:recipeqa_exp}
\end{figure*}

\begin{figure*}[!htb]
\minipage{0.45\textwidth}
  \includegraphics[width=\linewidth]{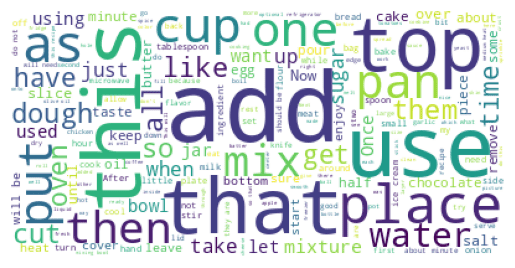}
  \subcaption{RecipeQA step text}
\endminipage\hfill
\minipage{0.45\textwidth}
  \includegraphics[width=\linewidth]{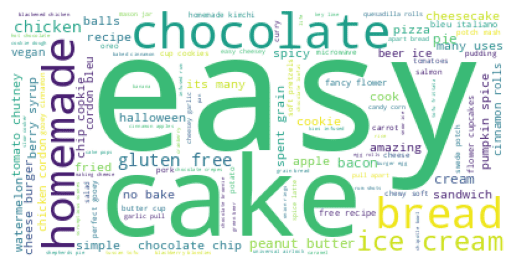}
  \subcaption{RecipeQA task title text}
\endminipage\hfill
\caption{Word cloud distributions of the task title and step text in the \recipe~ dataset.}
\label{fig:wordcloud_recipeqa}
\end{figure*}

\noindent\textbf{Dataset Statistics} 
We visualize two examples of our repurposed \recipe~ for multimodal procedural planning in Figure \ref{fig:recipeqa_exp}.
We also show the word-cloud distribution of task name and textual plan reference in Figure \ref{fig:wordcloud_recipeqa}.

\subsection{\wiki~}
\label{apped:wikihow}
\noindent\textbf{Raw Data Collection} To facilitate research on learning to generate procedural planning in a multimodal setting, we have constructed the large-scale \wiki~ dataset collected from the \wikihow~ website\footnote{\url{https://www.wikihow.com/}}, which is under an Attribution-Noncommercial-Share Alike 3.0 Creative Commons License.. This website provides a wide range of how-to articles related to everyday life topics, which are collaboratively written by its users and reviewed by experts.
We crawled each article, collecting the task title, URL, introduction, topics, and steps. Each step includes a brief textual action, a detailed context, and an illustration image. Our raw dataset consists of 30,026 examples across 19 categories and 2,062 topics. We plan to release the raw data in the hopes of pre-training models for procedural planning and knowledge reasoning.

\noindent\textbf{Quality Control} To improve the evaluation of different baselines, we further selected five categories that feature temporal actions and high-fidelity visual contexts: \textit{Food and Entertaining}, \textit{Hobbies and Crafts}, \textit{Home and Garden}, \textit{Pets and Animals}, and \textit{Sports and Fitness}. In order to reflect common tasks in real-life scenarios, articles with fewer than three steps or more than 22 steps were excluded, as well as articles with images of a dimension size of fewer than 400 pixels. Each category was balanced with 200 examples. To further ensure high quality, we conducted a quality control in which well-trained human annotators reviewed the dataset and manually revised the examples if there was any wrong or inappropriate content.

\begin{figure*}[!htb]
\minipage{0.99\textwidth}
\includegraphics[width=0.99\textwidth]{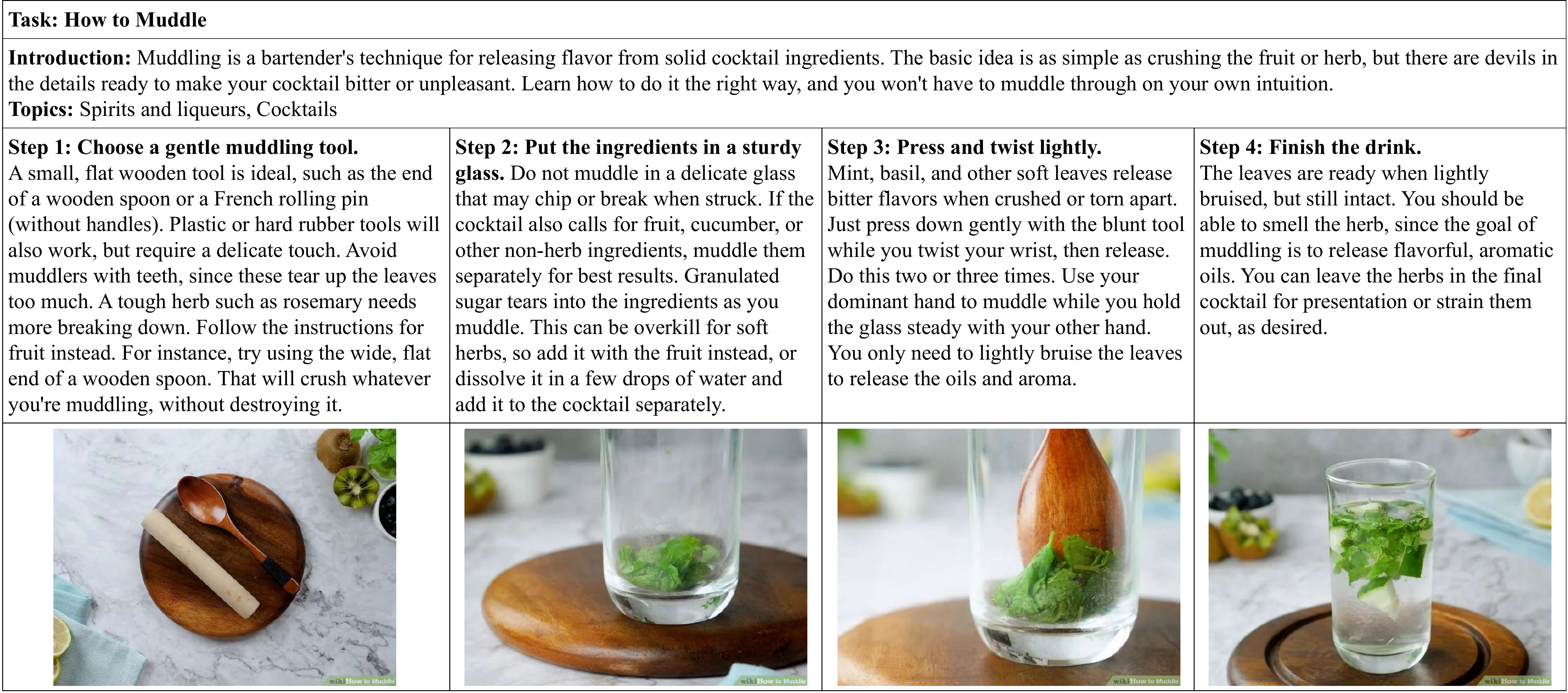}
\endminipage
\hfill
\vspace{5mm}
\minipage{0.99\textwidth}
\includegraphics[width=0.99\textwidth]{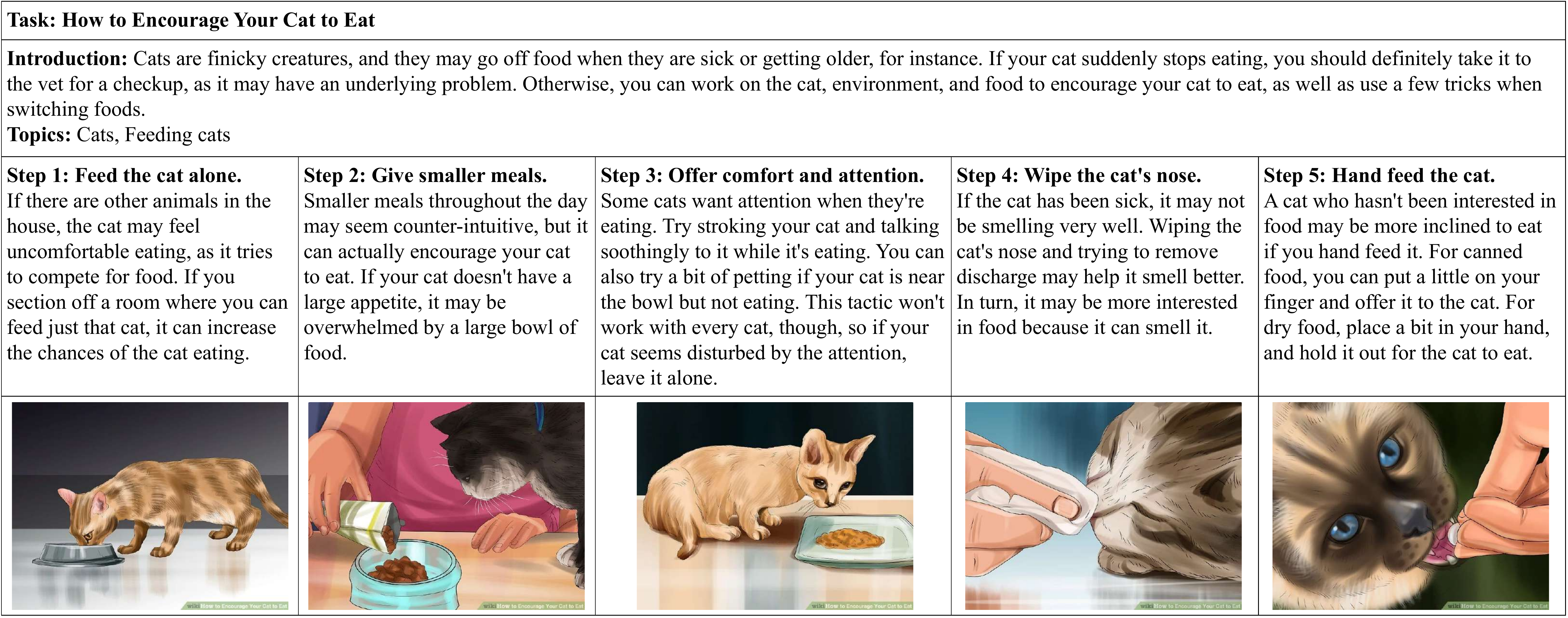}
\endminipage
\hfill
\vspace{5mm}
\minipage{0.99\textwidth}
\includegraphics[width=0.99\textwidth]{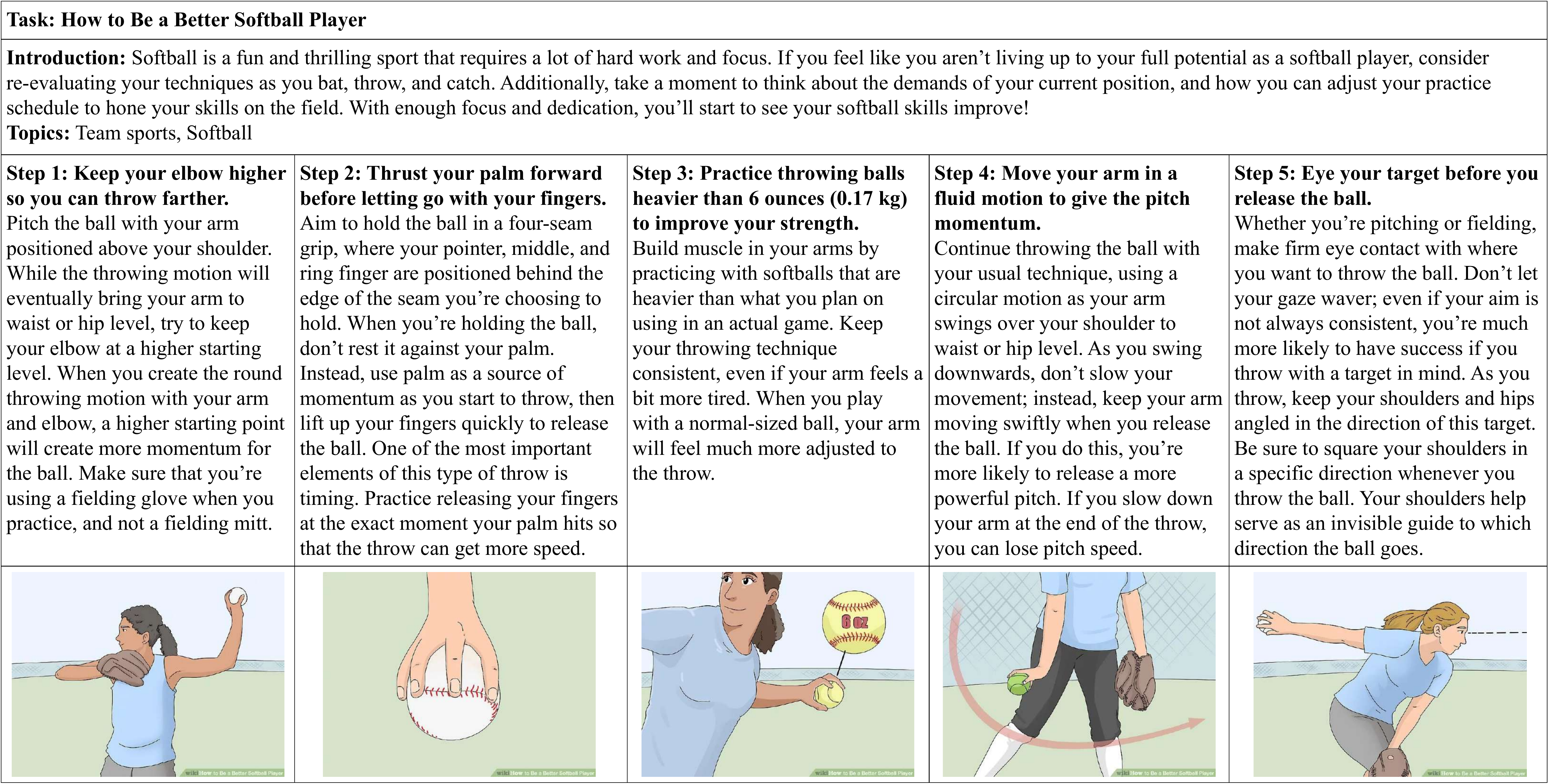}
\endminipage
\caption{Three examples in our curated \wiki~ dataset.}
\label{fig:wikihow_exp}
\end{figure*}

\begin{figure*}[!htb]
\minipage{0.45\textwidth}
  \includegraphics[width=\linewidth]{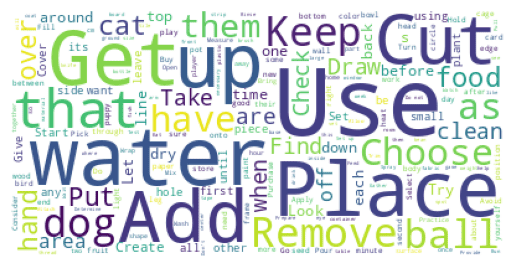}
  \subcaption{Wikihow step text}
\endminipage\hfill
\minipage{0.45\textwidth}
  \includegraphics[width=\linewidth]{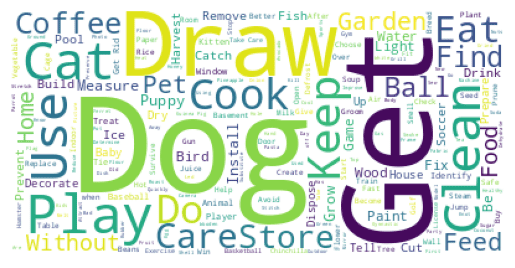}
  \subcaption{Wikihow task title text}
\endminipage\hfill
\caption{Word cloud distributions of the task title and step text in the \wiki~ dataset.}
\label{fig:wordcloud_wikihow}
\end{figure*}

\noindent\textbf{Dataset Statistics}  Finally, our \wiki~ dataset consists of 1,000 examples across 5 categories and 370 different topics. Three examples of the dataset are illustrated in Figure \ref{fig:wikihow_exp}, which include muddling mint leaves for a cocktail, encouraging a cat to eat, and becoming a better football player. Each example is composed of a title, introduction, related topics, and a list of detailed steps with visual aids. The word-cloud distributions of the task titles and step text are visualized in Figure \ref{fig:wordcloud_wikihow}, which demonstrate the diversity of topics and words. 

\section{Evaluation Details}
\subsection{Crowdsourcing Human Evaluation}
\label{app:amazon_mechanical_turk}
We manually ensure no personal information is collected and no offensive content is presented during human evaluations.
The hourly wage paid to participants is estimated at \$12. And the total amount spent on participant compensation is \$1958.

We average the results from $3$ annotators for each example.
Given the high-level goal (task name) for each assignment, we want the annotators to compare two generated text and image sequences in terms of \textit{Textual-Informativeness}, \textit{Visual-Informativeness}, \textit{Temporal Coherence} and \textit{Plan Accuracy}. Before going to the question, we let the annotators read the instructions below:

\textbf{Instruction:} Given the Task (e.g, Task: How to muddle), please compare two sequences of steps Sequence 1 and Sequence 2, and determine which sequence is better in terms of four aspects:
\begin{itemize}
    \item Textual-Informativeness: whether the textual sequence (the sequence of texts) contains the amount of information needed to complete the task.
    \item Visual-Informativeness: whether the visual sequence (the sequence of images) contains the amount of information needed to complete the task.
    \item Temporal Coherence: whether the multimodal sequence (the paired sequence of texts and images) meets the temporal commonsense requirements, such as a step occurring before another step instead of after.
    \item Plan Accuracy: whether the multimodal sequence (the paired sequence of texts and images) can successfully complete the task.
\end{itemize}
To be concrete, the annotators were asked to choose one from the two sequences by \textit{1 - Sequence 1 is better}, \textit{2 - Tie}, and \textit{3 - Sequence 2 is better}.
We provide the multimodal plans as follows:

\textbf{Task: How to Get Kids to Eat Healthy.}

\textbf{Sequence 1:}

Step 1: Talk to your kids about the importance of eating healthy and make sure that nutritious food is accessible and visible in the house.

Visual Plan at Step 1: [Image]

Step 2: Set rules or guidelines for what is allowed and not allowed in terms of snacks and meals., involve children in grocery shopping and meal preparation as much as you can, lead by example by practicing healthy eating habits yourself.

Visual Plan at Step 2: [Image]

Step 8: Encourage better dietary decisions at mealtimes by sitting down together with them at the table filled with various types of wholesome foods.

Visual Plan at Step 8: [Image]

\textbf{Sequence 2:}

Step 1: Talk to your kids about the importance of eating healthy.

Visual Plan at Step 1: [Image]

Step 2: Set rules or guidelines for what is allowed and not allowed in terms of snacks and meals.

Visual Plan at Step 2: [Image]

Step 8: Reward your child for trying new foods or making healthier choices.

Visual Plan at Step 8: [Image]

We show the paired visual plan at each step below the text plan.
\begin{figure*}[!htb]
\minipage{\textwidth}
  \includegraphics[width=\textwidth]{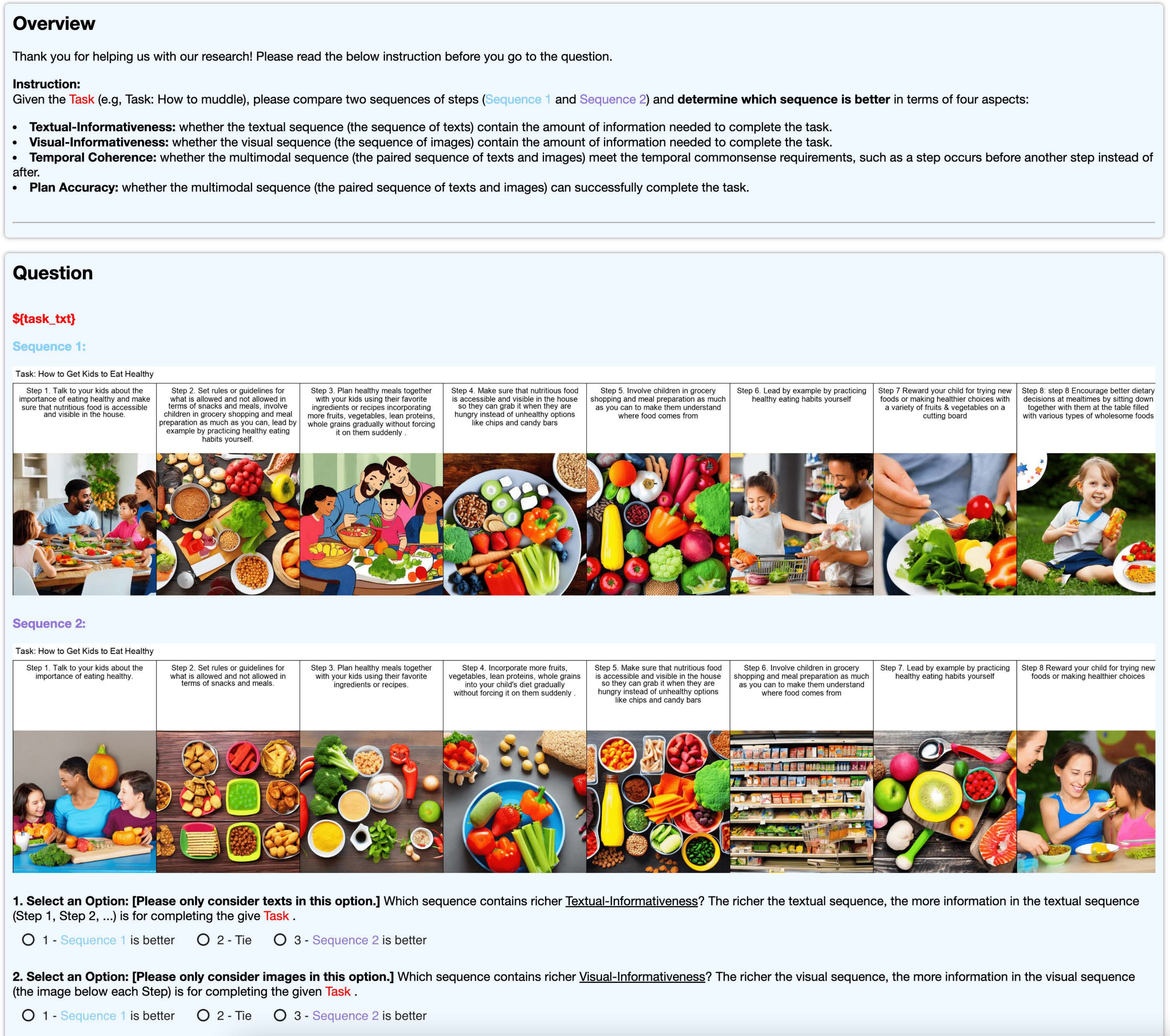}
\endminipage\hfill
\minipage{\textwidth}
  \includegraphics[width=\textwidth]{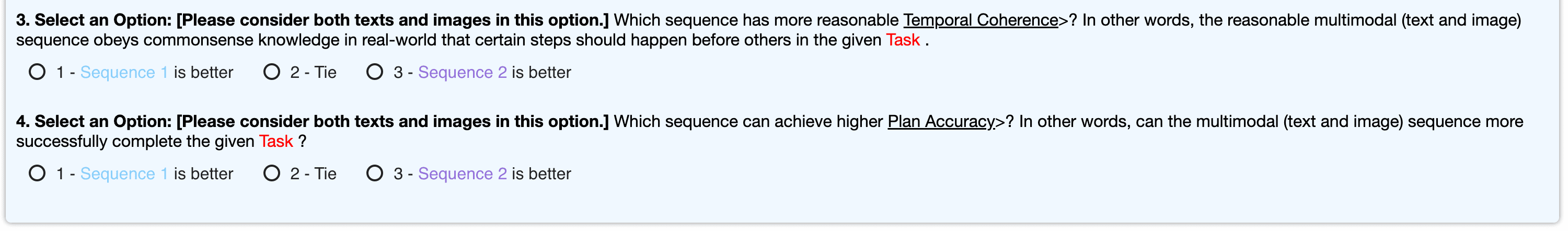}
\endminipage\hfill
\caption{Amazon Mechanical Turk Platform. Questions Layout for Human Raters for Win-Tie-Lose Comparison on \wiki.}\label{fig:mturk_head2head_wikihow}
\end{figure*}

\begin{figure*}[!htb]
\minipage{\textwidth}
  \includegraphics[width=\textwidth]{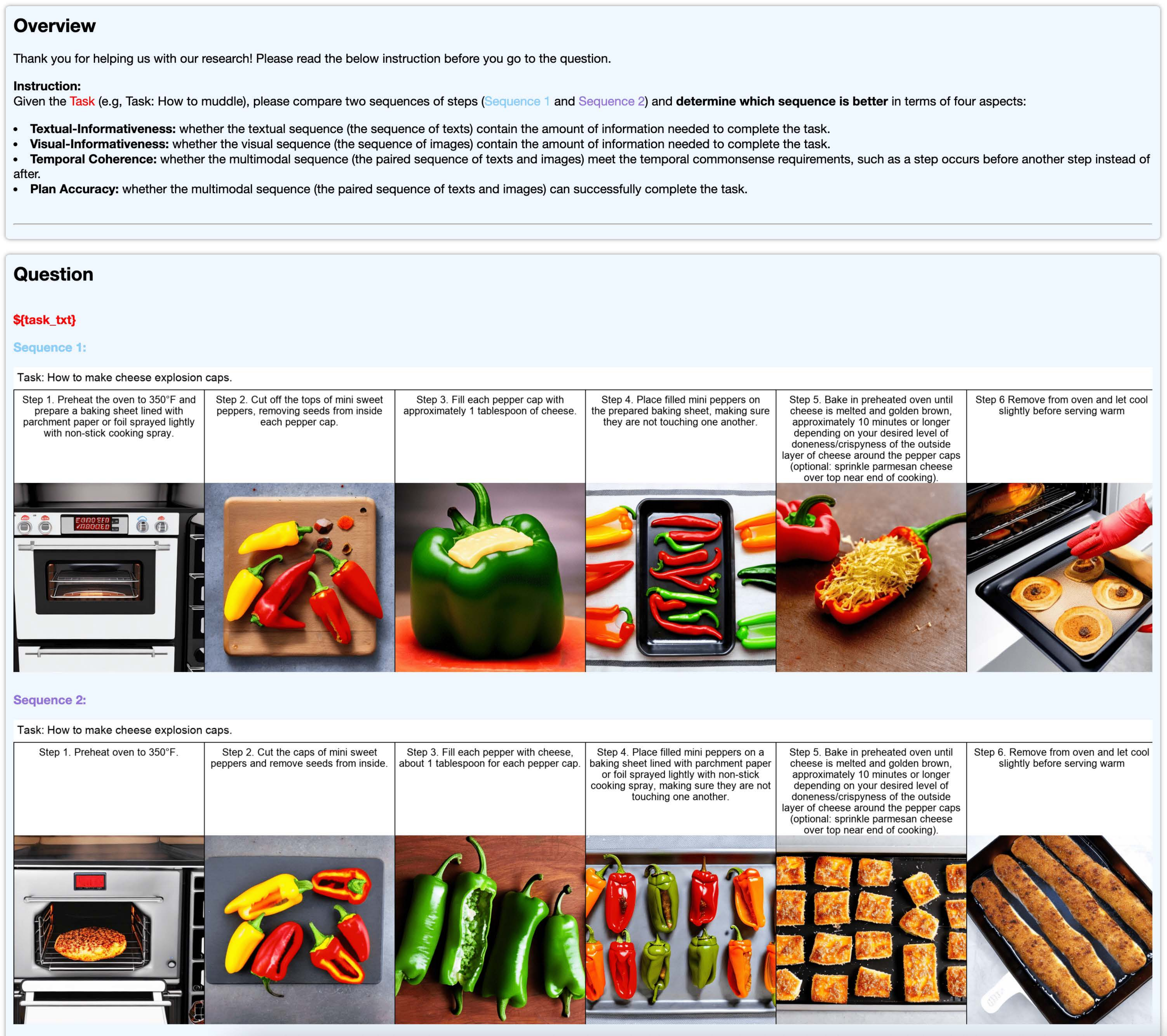}
\endminipage\hfill
\minipage{\textwidth}
  \includegraphics[width=\textwidth]{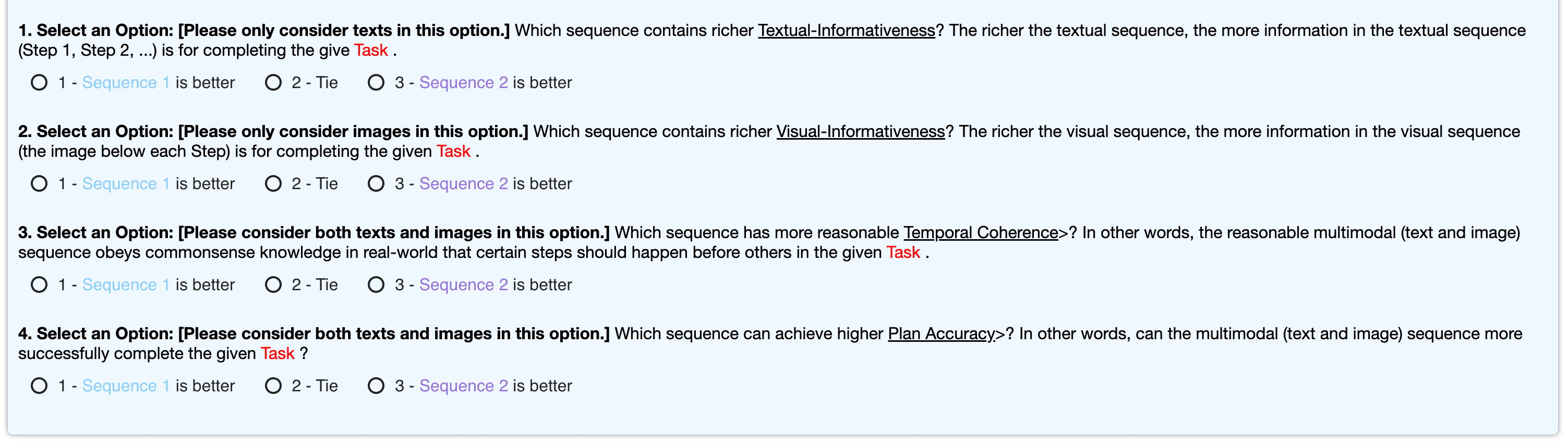}
\endminipage\hfill
\caption{Amazon Mechanical Turk Platform. Questions Layout for Human Raters for Win-Tie-Lose Comparison on \recipe~.}\label{fig:mturk_head2head_recipeqa}
\end{figure*}

Please refer to our Amazon Mechanical Turk human evaluation interface for head-to-head comparison on \wiki~ and \recipe~ in Figure~\ref{fig:mturk_head2head_recipeqa} and Figure~\ref{fig:mturk_head2head_wikihow} respectively.

\section{More Results}
\label{app:more_results}
\subsection{Showcases}
\label{app:showcases}

\begin{figure*}[!htb]
\minipage{\textwidth}%
  \includegraphics[width=\linewidth]{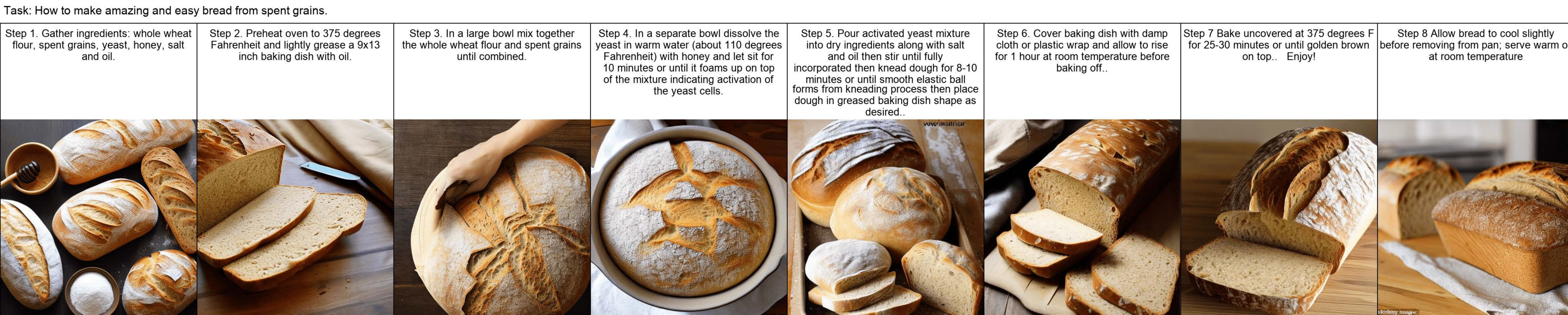}
\endminipage\hfill
\minipage{\textwidth}%
  \includegraphics[width=\linewidth]{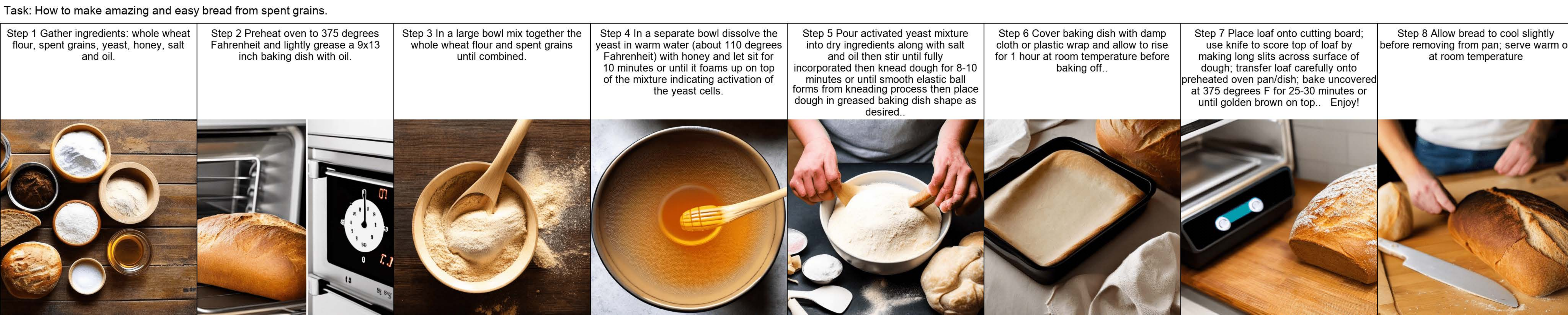}
  \subcaption{\uthree~ (Top) vs. \ours~ (Bottom)}
\endminipage\hfill
\minipage{\textwidth}%
  \includegraphics[width=\linewidth]{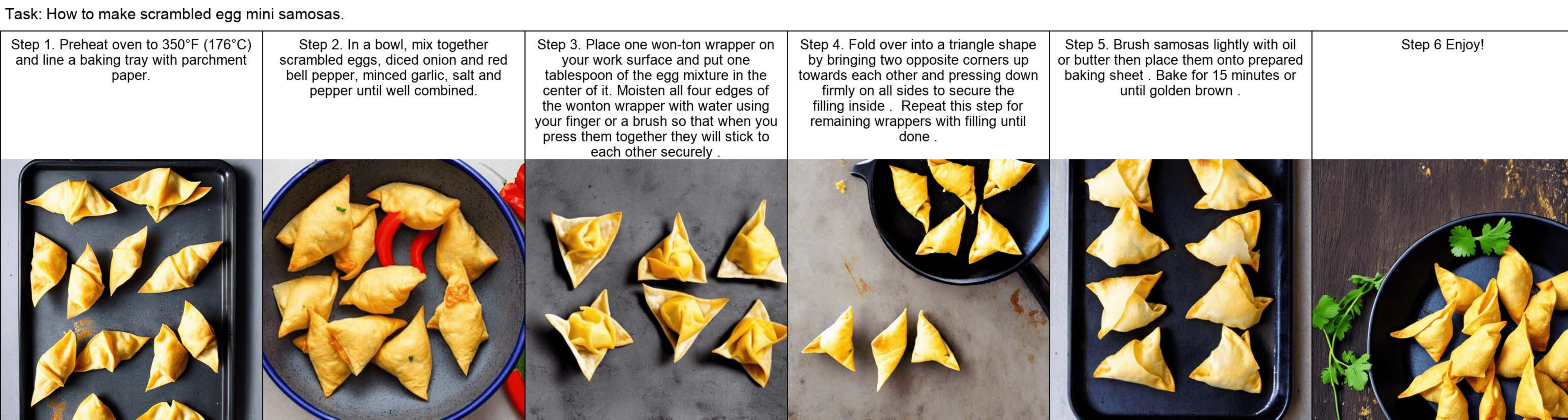}
\endminipage\hfill
\minipage{\textwidth}%
  \includegraphics[width=\linewidth]{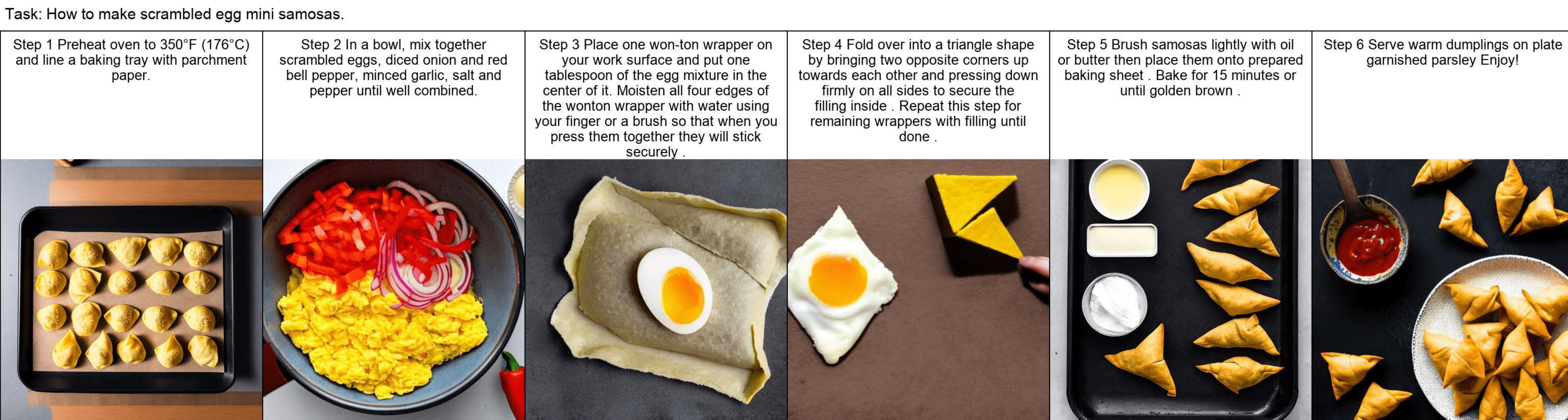}
  \subcaption{\uthree~ (Top) vs. \ours~ (Bottom)}
\endminipage\hfill
\caption{More showcases of plan comparisons on \recipe~.
}\label{fig:more_cases_1}
\end{figure*}

\begin{figure*}[!htb]

\minipage{\textwidth}%
  \includegraphics[width=\linewidth]{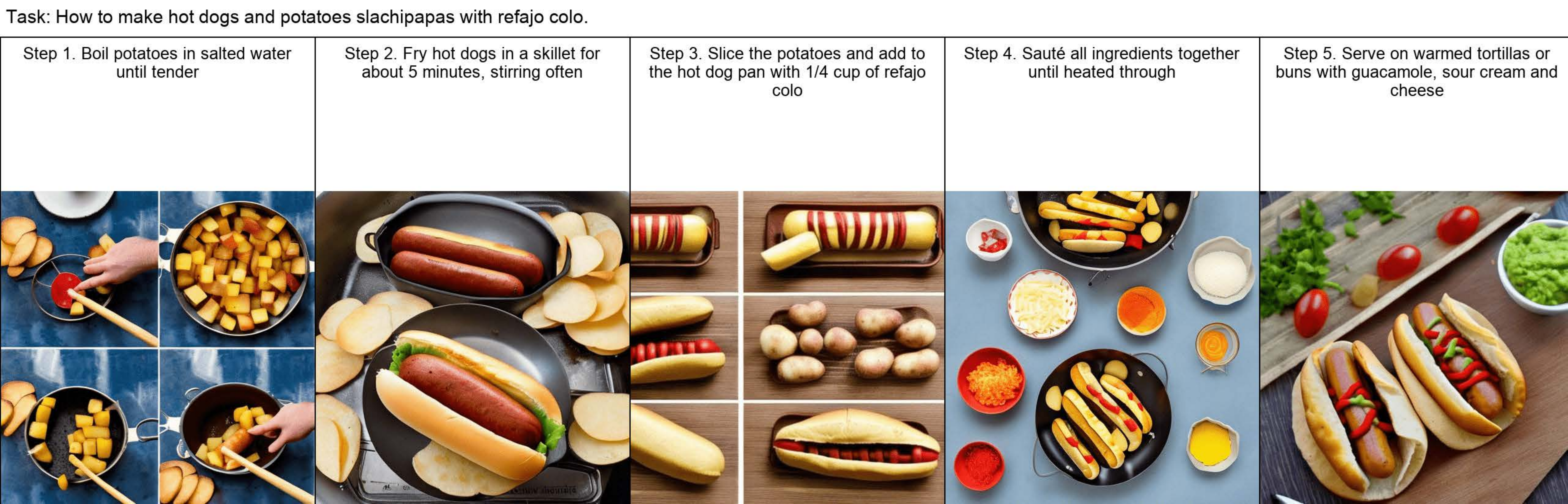}
\endminipage\hfill
\minipage{\textwidth}%
  \includegraphics[width=\linewidth]{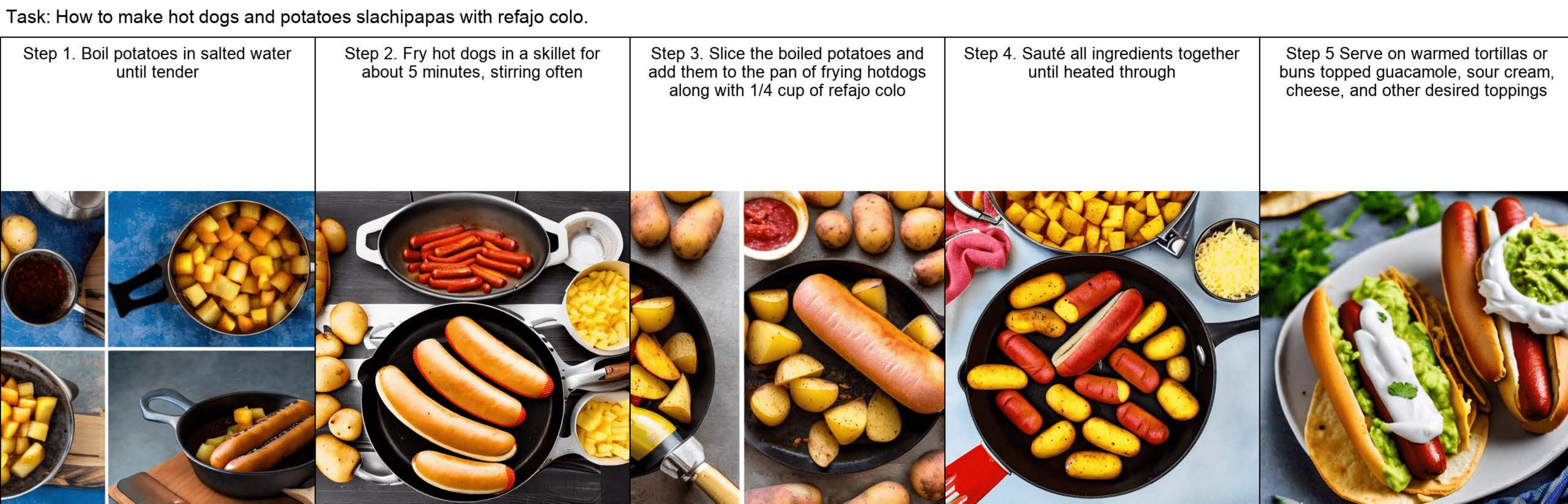}
  \subcaption{\uthree~ (Top) vs. \ours~ (Bottom)}
\endminipage\hfill
\minipage{\textwidth}%
  \includegraphics[width=\linewidth]{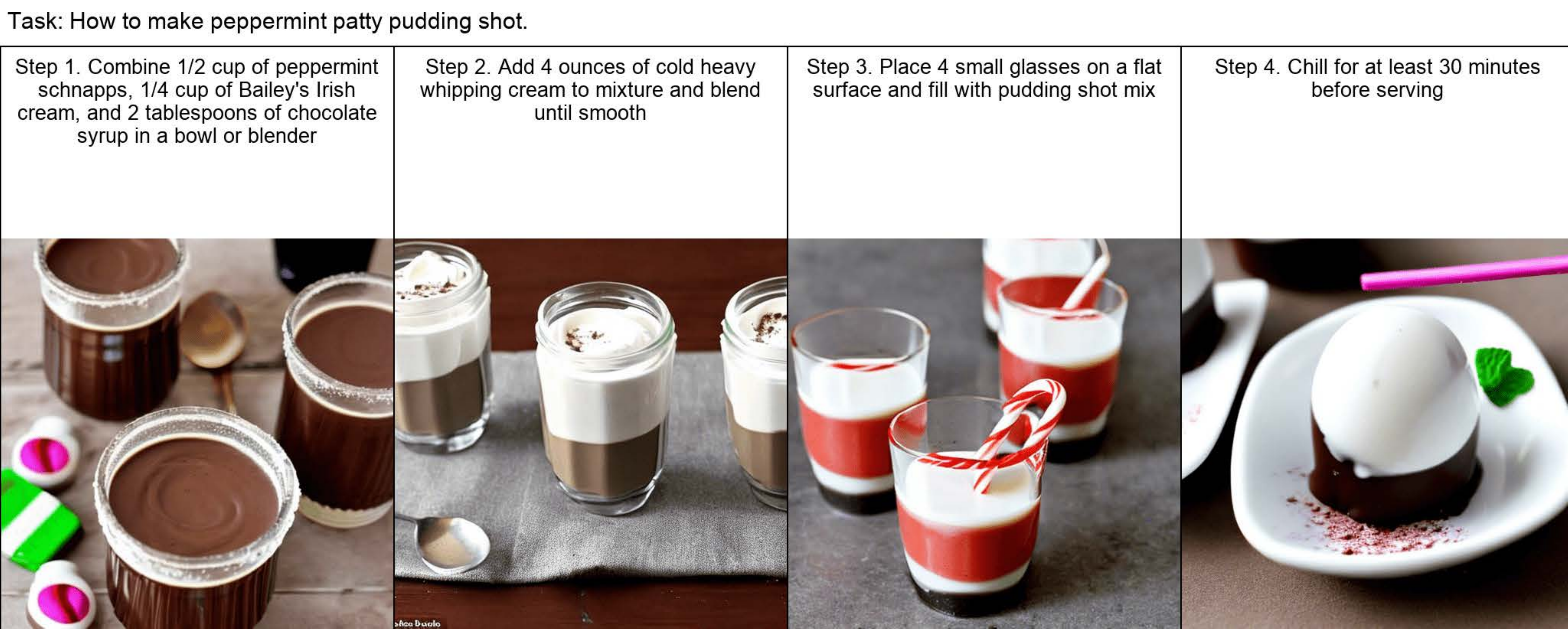}
\endminipage\hfill
\minipage{\textwidth}%
  \includegraphics[width=\linewidth]{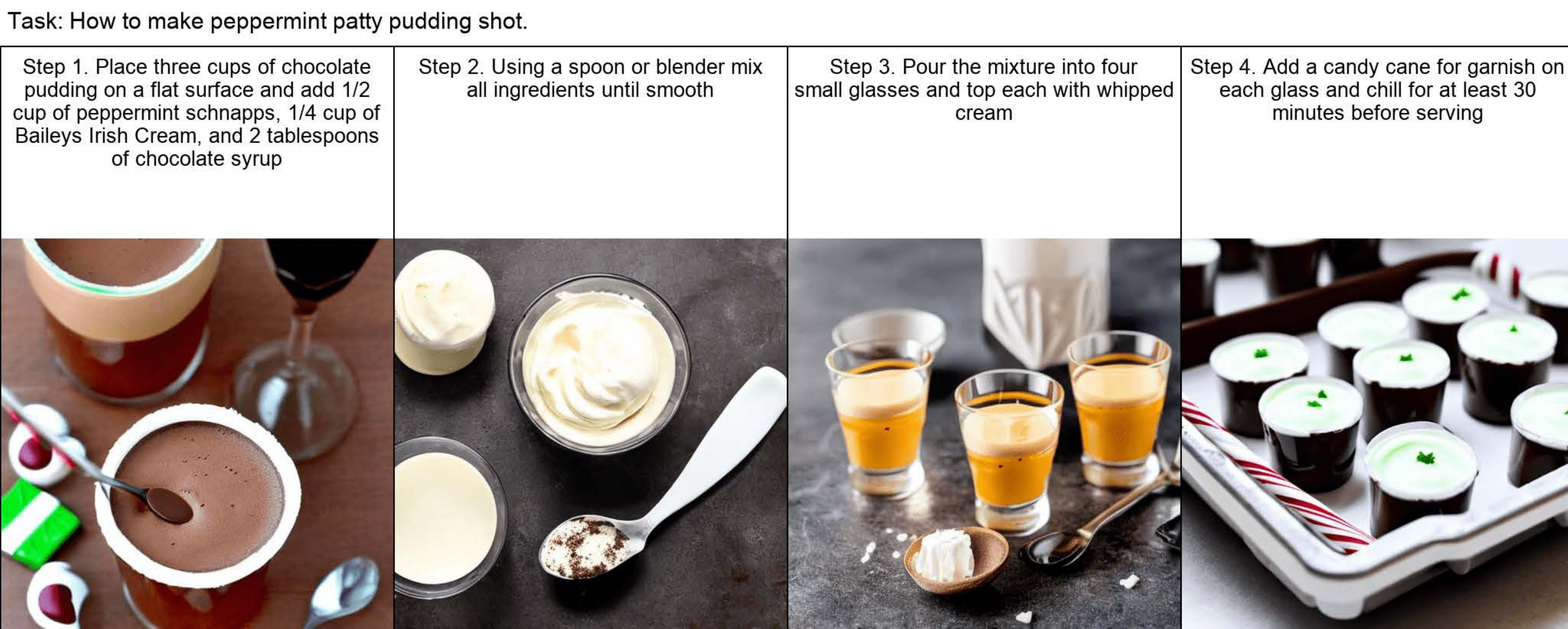}
  \subcaption{\uthree~ (Top) vs. \ours~ (Bottom)}
\endminipage\hfill
\caption{More showcases of plan comparisons on \recipe~.
}\label{fig:more_cases_2}
\end{figure*}

\begin{figure*}[!htb]

\minipage{\textwidth}%
  \includegraphics[width=\linewidth]{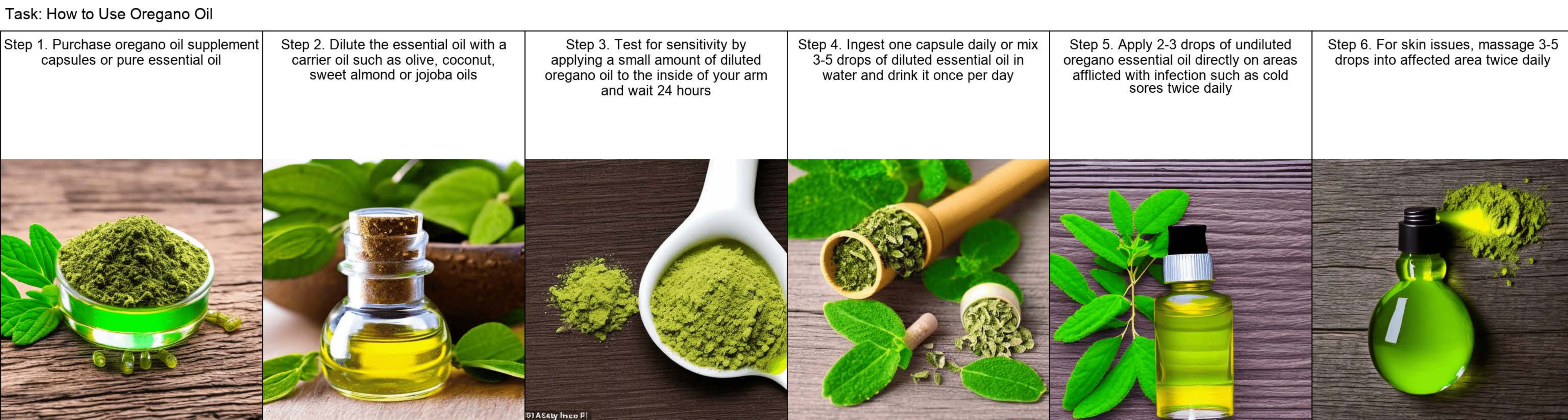}
\endminipage\hfill
\minipage{\textwidth}%
  \includegraphics[width=\linewidth]{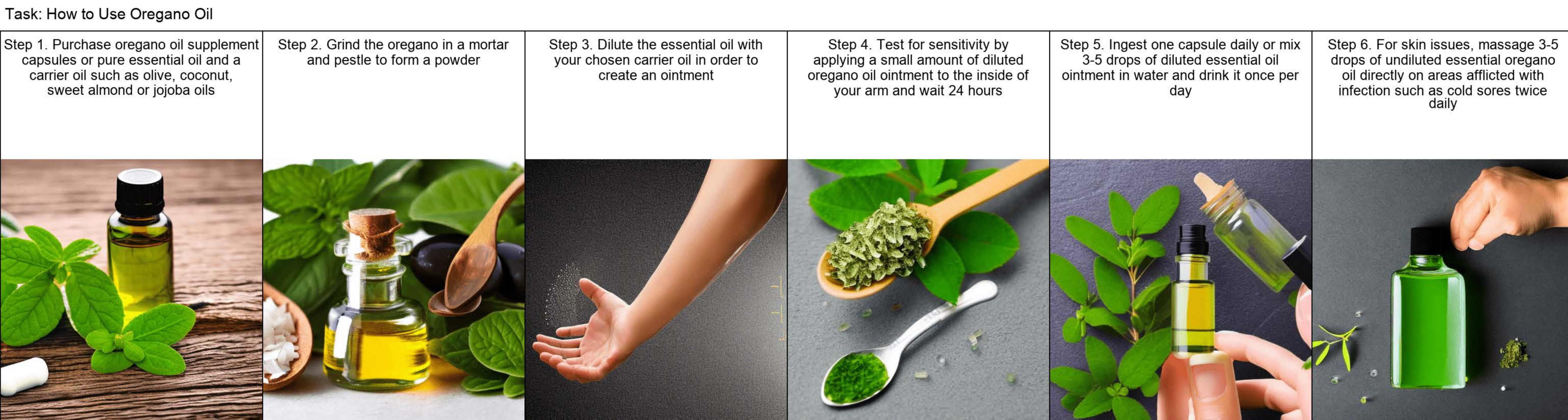}
  \subcaption{\uthree~ (Top) vs. \ours~ (Bottom)}
\endminipage\hfill
\minipage{\textwidth}%
  \includegraphics[width=\linewidth]{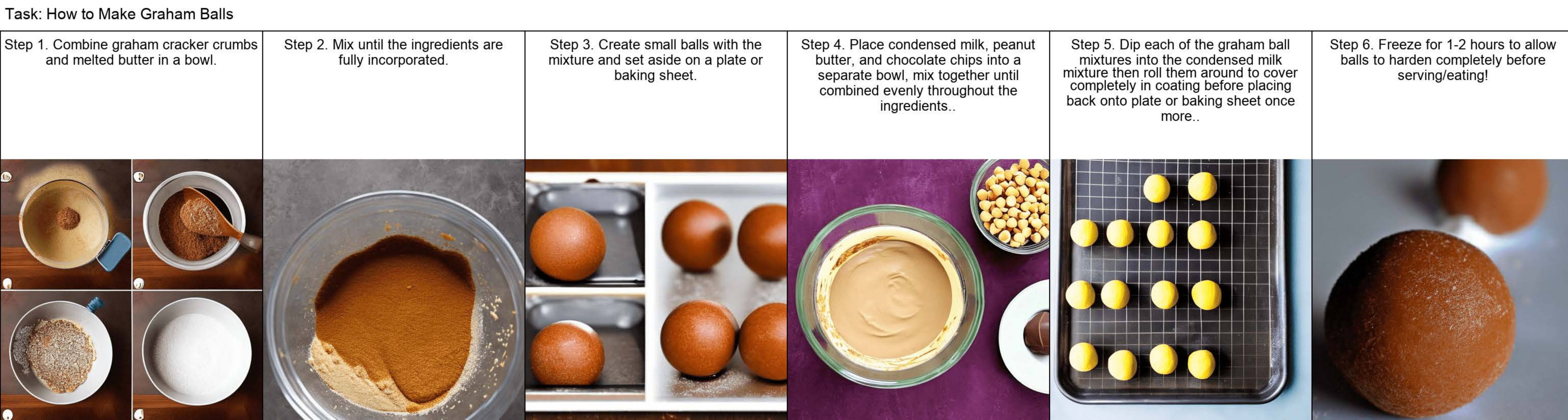}
\endminipage\hfill
\minipage{\textwidth}%
  \includegraphics[width=\linewidth]{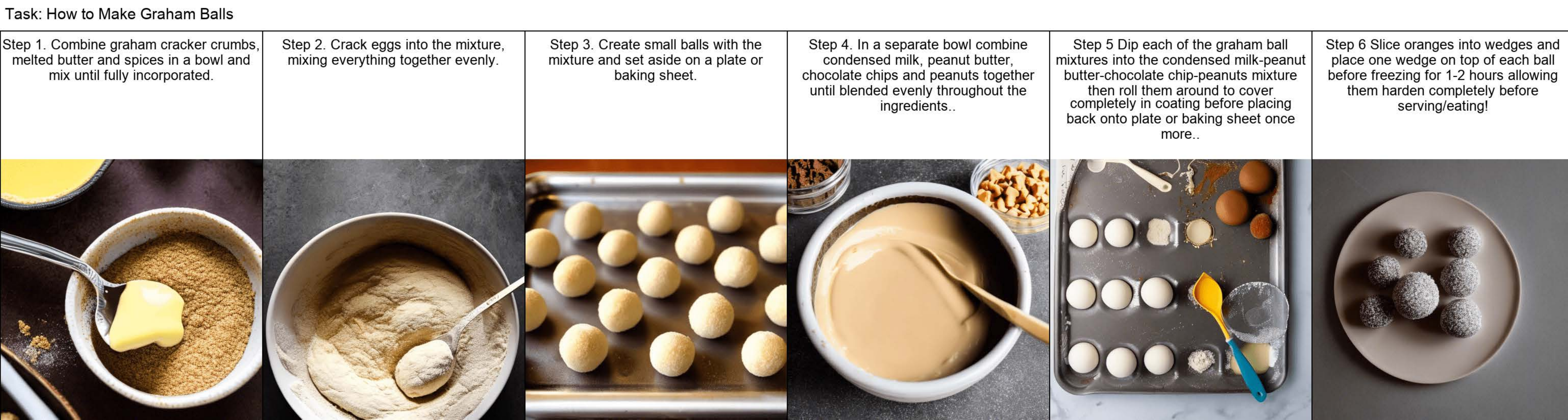}
  \subcaption{\uthree~ (Top) vs. \ours~ (Bottom)}
\endminipage\hfill
\caption{More showcases of plan comparisons on \wiki~.
}\label{fig:more_cases_3}
\end{figure*}

\begin{figure*}[!htb]

\minipage{\textwidth}%
  \includegraphics[width=\linewidth]{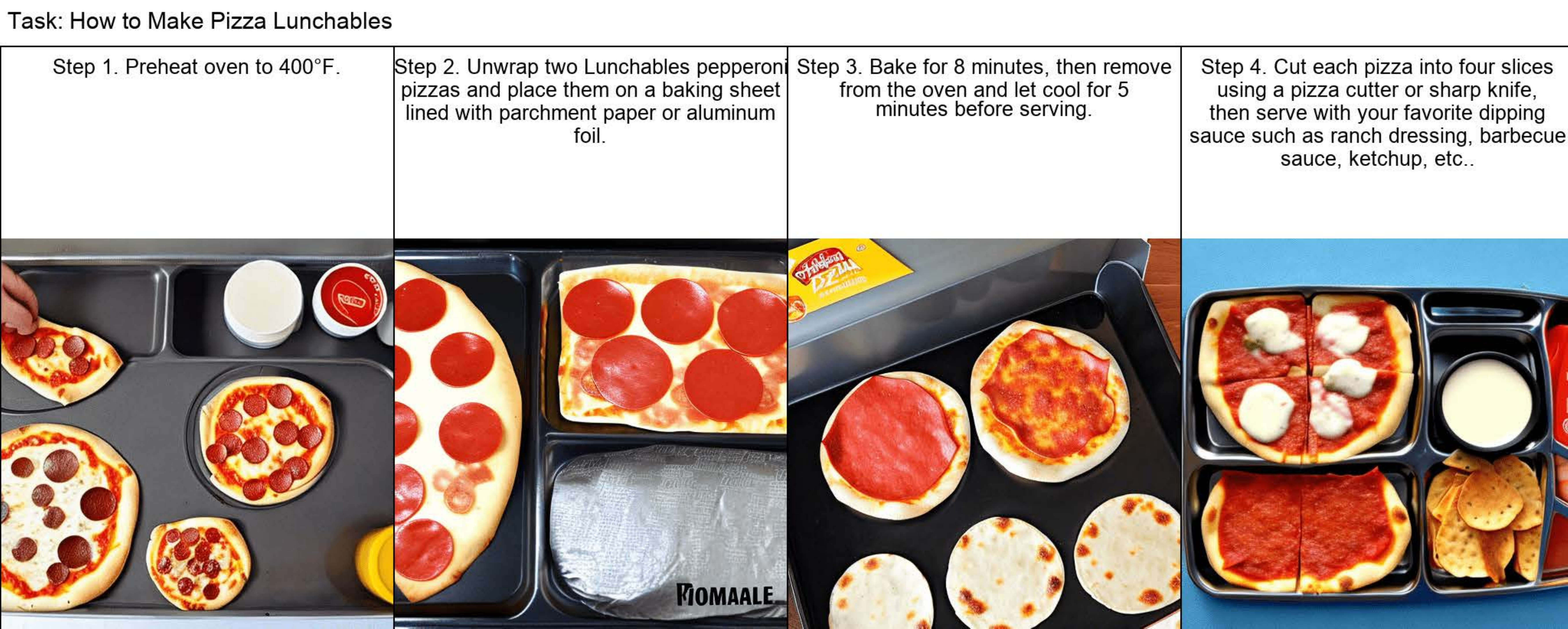}
\endminipage\hfill
\minipage{\textwidth}%
  \includegraphics[width=\linewidth]{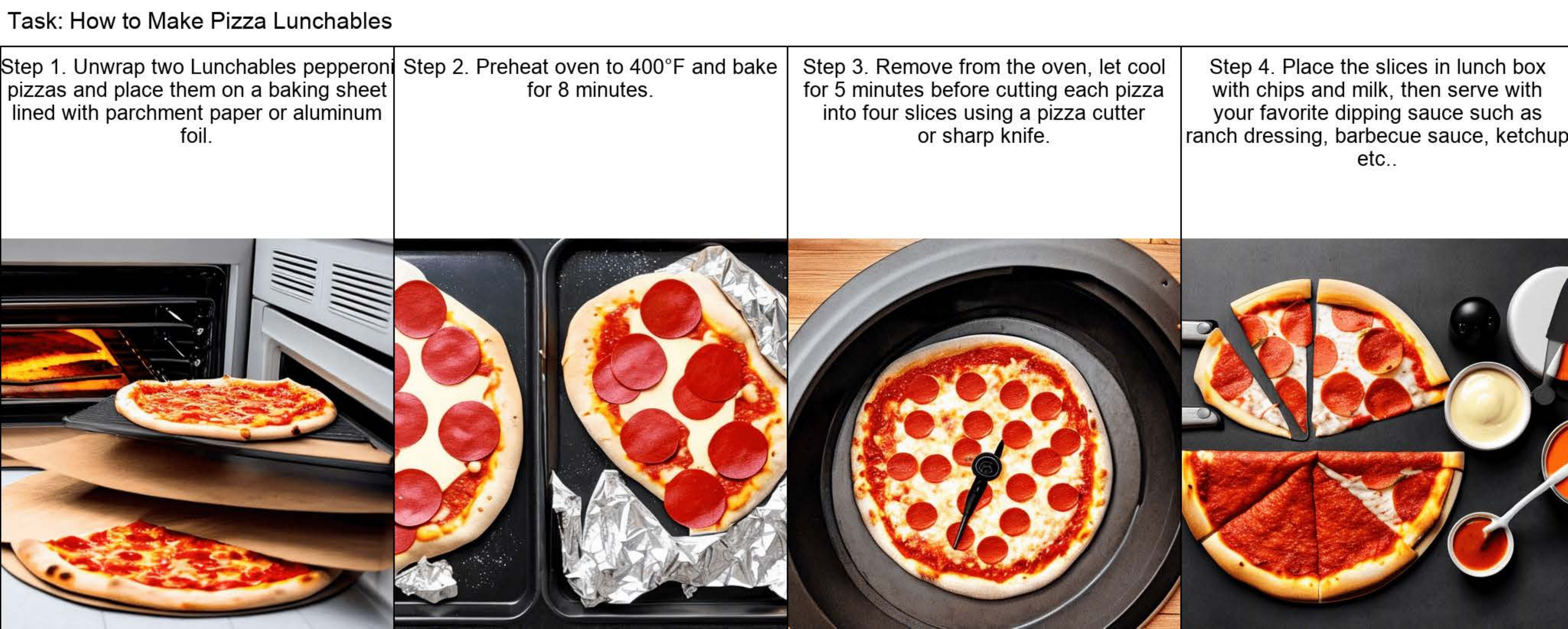}
  \subcaption{\uthree~ (Top) vs. \ours~ (Bottom)}
\endminipage\hfill
\minipage{\textwidth}%
  \includegraphics[width=\linewidth]{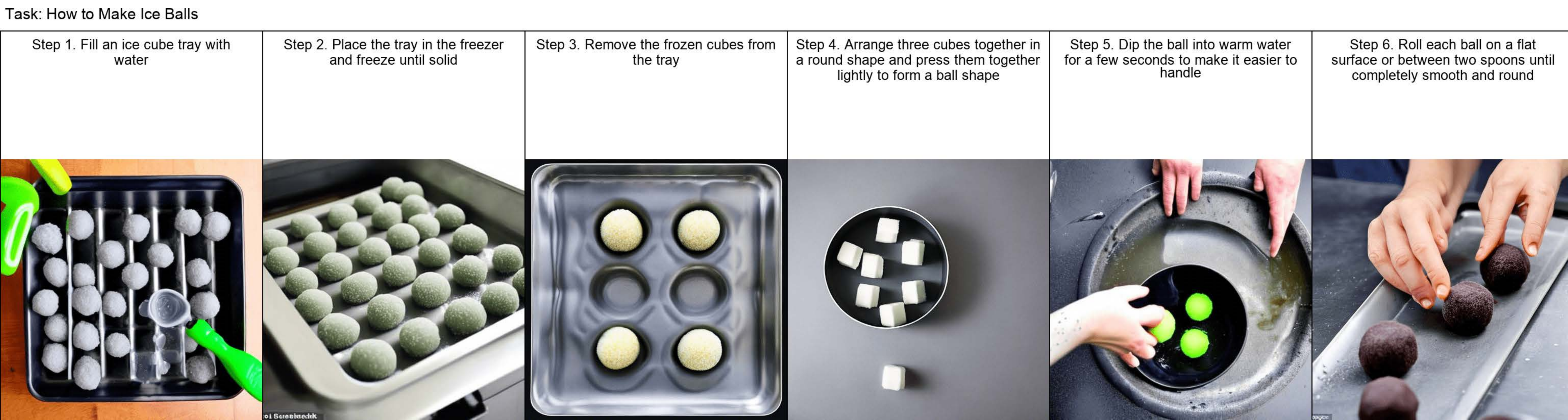}
\endminipage\hfill
\minipage{\textwidth}%
  \includegraphics[width=\linewidth]{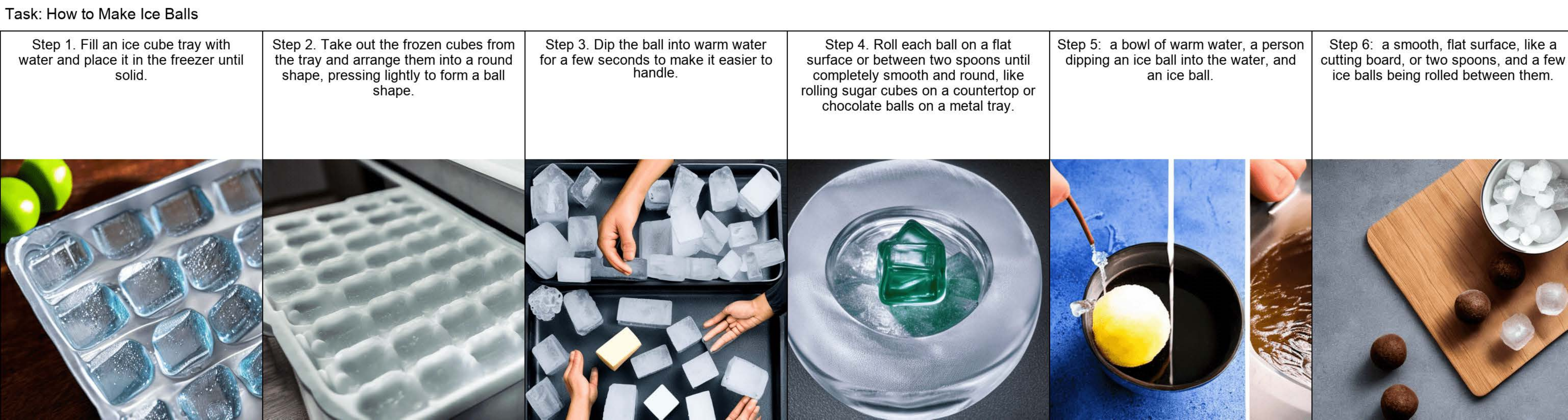}
  \subcaption{\uthree~ (Top) vs. \ours~ (Bottom)}
\endminipage\hfill
\caption{More showcases of plan comparisons on \wiki~.
}\label{fig:more_cases_4}
\end{figure*}

We show more cases in Figure~\ref{fig:more_cases_1}-~\ref{fig:more_cases_4} comparing our \method~ with powerful baselines \uthree~.

\subsection{Word Cloud}
In comparison with the word cloud distribution of the ground truth, we also show the word cloud of the baselines and \ours~ on \wiki~ and \recipe~.

\subsection{Failure Cases}
\begin{figure*}[t!]
\minipage{\textwidth}%
  \includegraphics[width=\linewidth]{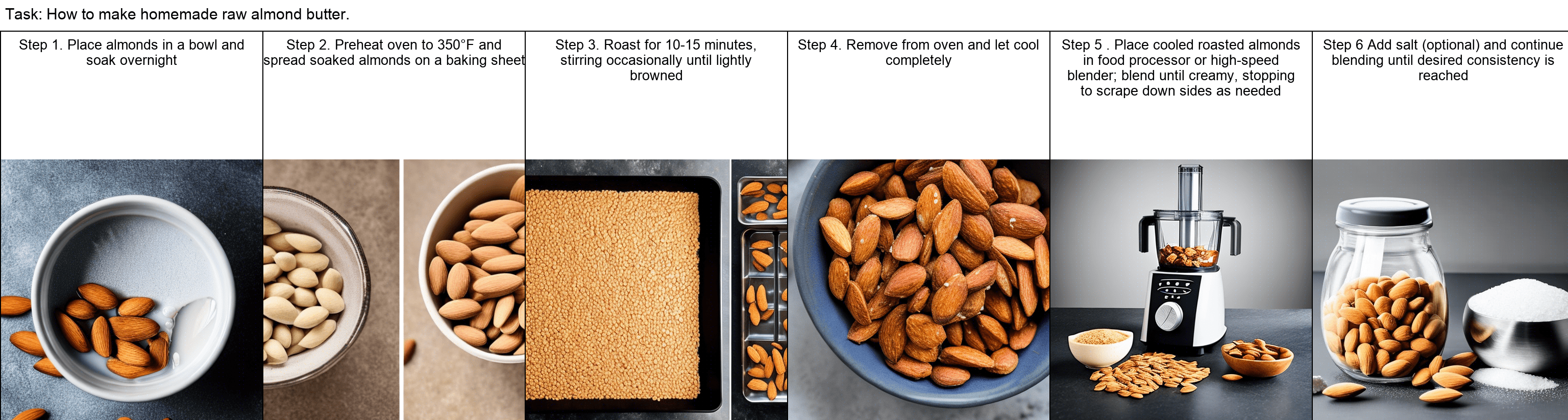}
  \subcaption{Failure multimodal plans generated by our \method~ (\ours~).}\label{fig:failure_cases_1}
\endminipage\hfill
\vspace{1mm}
\minipage{\textwidth}%
  \includegraphics[width=\linewidth]{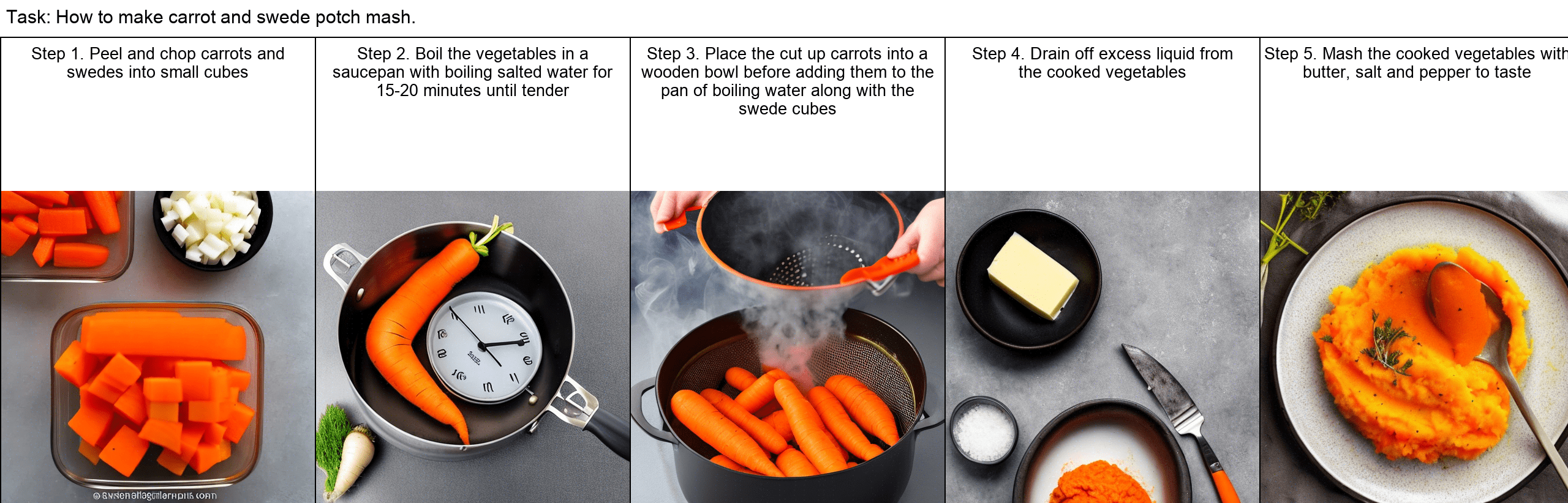}
  \subcaption{Failure multimodal plans generated by our \method~ (\ours~).}\label{fig:failure_cases_2}
\endminipage\hfill
\caption{We showcase failure cases of our \method~ on generating multimodal plans on both datasets.
}\label{fig:failure_cases}
\end{figure*}

In Figure~\ref{fig:failure_cases}, we showcase failure generation. For example, the state of the almond stays unchanged in Figure~\ref{fig:failure_cases_1}, we suppose this is due to no explicit awareness of previous state change.
In Figure~\ref{fig:failure_cases_2}, at step 2, the generated image plan, though complemented with the text plan, loses authenticity in that the clock should not appear in a pan with carrots.

\end{document}